\documentclass{article}
\usepackage{iclr2027_conference,times}
\iclrfinalcopy

\usepackage{microtype}
\usepackage{amsmath,amssymb}
\usepackage{natbib}
\usepackage{hyperref}
\usepackage{url}
\usepackage{graphicx}
\usepackage{booktabs}
\usepackage{multirow}
\usepackage{colortbl,xcolor}
\usepackage{wrapfig}
\usepackage{subcaption}
\usepackage{pifont}
\usepackage[most]{tcolorbox}
\tcbuselibrary{skins,breakable}
\usepackage[table]{xcolor}
\usepackage{minitoc}

\newcommand{\xmark}{\ding{55}}
\definecolor{GraphColor}{HTML}{677a98}
\definecolor{SemanticColor}{HTML}{6e5bd9}
\definecolor{GraphSemanticColor}{HTML}{3d99e4}
\definecolor{LLMColor}{HTML}{e05e8a}
\definecolor{OursColor}{HTML}{e99043}

\definecolor{promptbox_color}{HTML}{f8dbc2}
\definecolor{darkblue}{rgb}{0,0,0.5}
\hypersetup{colorlinks=true,citecolor=darkblue,linkcolor=darkblue,urlcolor=darkblue}

\newcommand{\revision}[1]{\textcolor{black}{#1}}

\title{LLM4Impact: Integrating Heterogeneous \\ Information for Scientific Impact Prediction}

\author{Yong Cao\textsuperscript{1}, Markus Flicke\textsuperscript{1}, Haoyu He\textsuperscript{1}, Katrin Renz\textsuperscript{1}, Andreas Geiger\textsuperscript{1, 2}
\\
\textsuperscript{1}University of Tübingen, Tübingen AI Center \quad
\textsuperscript{2}KE:SAI
\\
\small{
\textbf{Website:} \href{https://yongcaoplus.github.io/llm4impact/}{yongcaoplus.github.io/llm4impact} 
} 
\small{
\textbf{Code:} \href{https://github.com/yongcaoplus/llm4impact}{github.com/yongcaoplus/llm4impact}
}
}

\begin{document}
\doparttoc 
\faketableofcontents 

\maketitle

\vspace{-10mm}
\begin{figure}[h]
    \centering
    \includegraphics[width=0.93\textwidth]{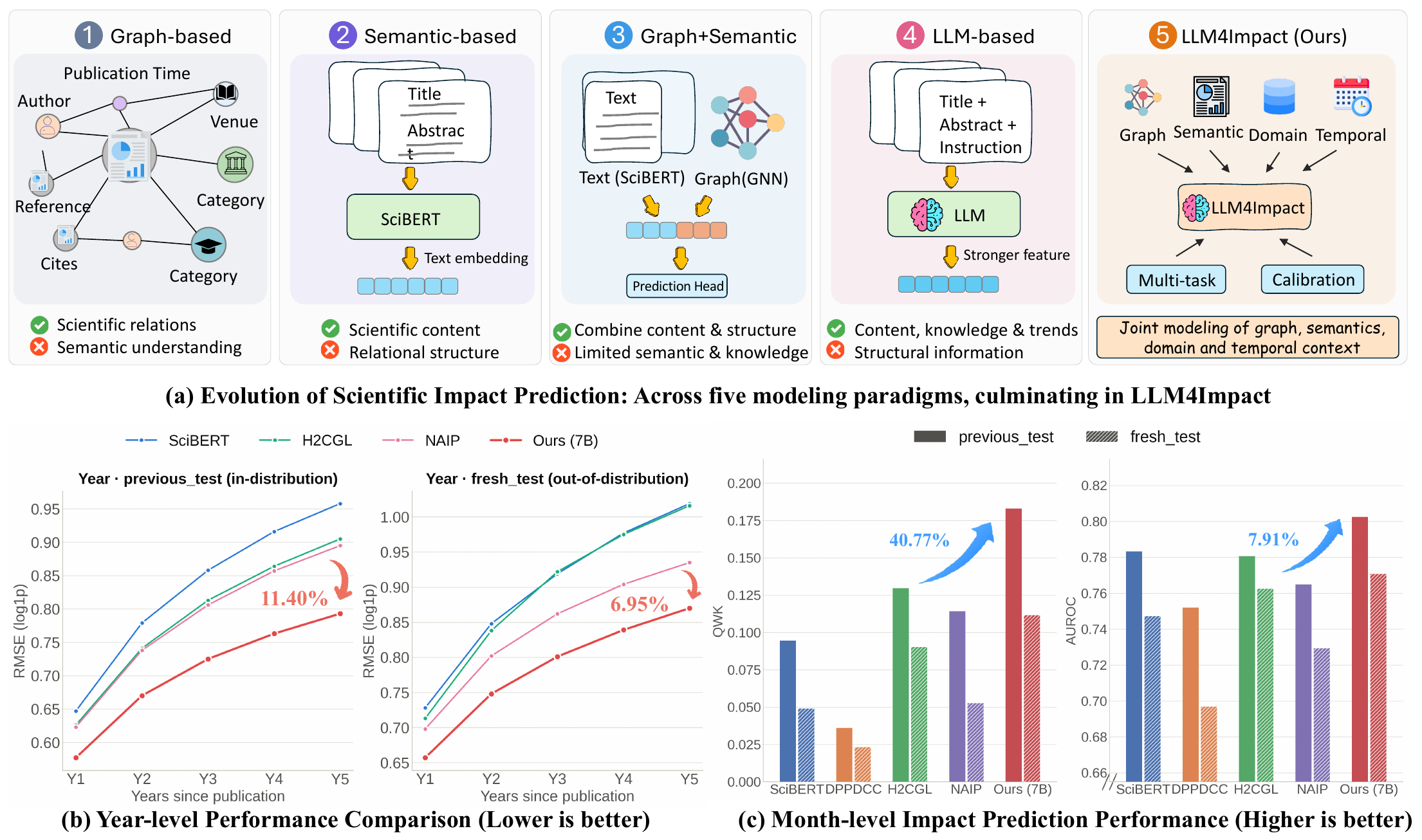}
    \caption{\textbf{Overview of LLM4Impact and its performance.}
(a) Scientific impact prediction across five paradigms:
\textcolor{GraphColor}{graph-based}, 
\textcolor{SemanticColor}{semantic-based},
\textcolor{GraphSemanticColor}{graph+semantic},
\textcolor{LLMColor}{LLM-based},
and \textcolor{OursColor}{LLM4Impact}, which integrates graph, semantic, domain, and temporal information with LLMs; (b--c)  Performance on both previous and fresh test sets demonstrates consistent gains across prediction settings.}
    \label{fig:teaser_fig}
\end{figure}

\begin{abstract}
Predicting the future impact of a newly published paper is challenging because it must be inferred from heterogeneous evidence available at publication time. Existing approaches often rely on a single source of information or combine multiple sources without accounting for their different predictive roles. In this paper, we present \emph{LLM4Impact}, an evidence-aware method for scientific impact prediction that learns to represent, integrate, and calibrate heterogeneous information. LLM4Impact combines semantic, graph, LLM, and temporal representations, and injects graph information into a frozen LLM through continuous prefix tokens. A context aware gating mechanism adaptively weights different evidence, while a separate calibration module accounts for domain and temporal variation in citation scales.  We further construct a large-scale benchmark dataset with 2 million papers, leakage-safe point-in-time heterogeneous ego graphs, temporal splits, and both year-level and month-level citation targets. Experiments show that LLM4Impact consistently outperforms strong semantic, graph, and LLM based baselines, with a 10.13\% reduction in year RMSE on the in distribution test set and a 6.87\% reduction under out-of-domain distribution. Our results reveal that the value of such evidence is context dependent: different papers benefit from different sources, while domain and publication time affect how evidence translates into citations. This finding motivates adaptive evidence selection and context-conditioned calibration rather than simply richer representations.
We will release our code, benchmark, and an interactive website demonstration upon publication.
\end{abstract}

\section{Introduction}

Predicting the future impact of scientific research is a fundamental problem in the science of science, with applications in identifying emerging research and supporting research evaluation \citep{bai_measure_2020, xu2022heterogeneous, xia2023review, zhang2024predicting}. A particularly challenging setting is cold start prediction, where a newly published paper has no citation history and its future impact must be inferred from information available at publication time. As shown in Figure~\ref{fig:teaser_fig}(a), existing approaches have explored semantic content \citep{beltagy-etal-2019-scibert, cohan2020specter}, scholarly structure \citep{he_h2cgl_2023, xue2024predicting}, temporal information \citep{wang2013quantifying, jiang2021hints, holm_longitudinal_2021}, and more recently LLMs \citep{zhao2025words, de2024can}. These signals are complementary, but their predictive value can vary across papers, domains, and publication periods. Effective cold start prediction therefore requires not only representing heterogeneous evidence, but also determining when and how each source should be used.


We identify three challenges that reveal why simply adding more information is insufficient.
First, publication time can fall outside the range observed during training.
A model based on discrete year embeddings cannot represent unseen publication years without learned embeddings, making temporal generalization difficult.
Second, the predictive value of semantic, graph, and LLM evidence is instance dependent:
sources that are informative for one paper may be less useful for another, making fixed or naive fusion suboptimal.
Third, domain and publication time affect how evidence maps to citation outcomes, since citation scales vary systematically across contexts.


To address these challenges, we propose \emph{LLM4Impact}, an
evidence-aware method that jointly models semantic, heterogeneous graph,
and LLM evidence under domain and temporal context for cold-start scientific
impact prediction. These sources provide complementary signals: semantic
representations capture paper content, heterogeneous graphs provide
relational evidence, and LLM representations capture richer contextual
signals. LLM4Impact preserves them as separate branches and injects graph
information into the LLM through continuous prefix tokens, allowing
structural and textual evidence to interact.
Building on this multi-branch design, LLM4Impact introduces three components.
First, a diagnostic-driven temporal encoder represents publication time with
a continuous function and a zero-initialized residual, enabling generalization
to unseen years. Second, a domain- and time-aware gated fusion module learns
instance-specific weights over the semantic, graph, and LLM branches,
reflecting their varying predictive value across contexts. Third, a decoupled
context-conditioned calibration module adjusts predictions for domain and
temporal variation in citation scales. Together, these components capture a
central insight: effective impact prediction requires not only richer
evidence, but also learning \emph{which evidence to use} and \emph{how to
interpret it in context}.


For extensive training and evaluation, we construct a large-scale benchmark from the Semantic Scholar database containing 2 million query papers published between 2005 and 2020. For each query paper, we construct a leakage safe heterogeneous ego graph using only information available at the paper's publication time. The benchmark contains both an in distribution test set and a temporally shifted test set consisting of papers from later publication years. We evaluate both five year citation prediction and fine grained monthly prediction. This setting allows us to examine not only prediction accuracy, but also the ability to generalize when the publication time is outside the training range.

As shown in Figure~\ref{fig:teaser_fig}(b-c), LLM4Impact consistently improves over strong semantic, graph based, and LLM based baselines. On the in distribution test set, five year RMSE drops from 0.790 to 0.710 (10.13\%), and under temporal distribution shift, from 0.844 to 0.786 (6.87\%). At the monthly level, LLM4Impact improves QWK by 40.77\% in distribution and 24.44\% under temporal distribution shift.
These results show that the benefit of heterogeneous information persists across prediction horizons and temporal shifts. More importantly, ablations reveal that the LLM, graph, and calibration components contribute complementary gains, supporting the view that effective impact prediction depends on both evidence selection and context-specific calibration. We also provide an interactive demo for counterfactual probing, allowing controlled changes to venue, authors, publication time, or references while holding other information fixed to examine changes in predicted impact. In summary, our main contributions are as follows:

\begin{itemize}

\item \textbf{A new large-scale benchmark dataset for scientific impact prediction.} We construct and release a 2-million-paper benchmark with leakage-safe point-in-time heterogeneous ego graphs, temporal train/test splits, and support - for the first time - both year-level and month-level citation targets, together with data construction pipeline for future research.

\item \textbf{An evidence-aware method for context-adaptive impact prediction.} We propose \emph{LLM4Impact}, which dynamically selects and calibrates heterogeneous evidence under temporal and domain context, enabling both robust prediction under distribution shift and controlled probing of how contextual factors affect scientific impact.


\item \textbf{Insights into heterogeneous evidence for impact prediction.}
Across prediction granularities and temporal shifts, controlled ablations,
pruning, and counterfactual probing reveal that evidence sources provide
complementary and context-dependent signals, while the interactive demo
enables controlled what-if analysis of how changes in authors, venues,
references, and other contextual factors affect predicted impact.

\end{itemize}
\section{Related Work}

\paragraph{Predicting Scientific Impact.}  Predicting scientific impact has long relied on citation counts as a proxy for influence \citep{fu2008models, vincent_relationship_2009, bai2019predicting}. Traditional bibliometrics like the h-index reflect established rather than future impact, prompting interest in article impact prediction based on early signals \citep{yang2023revisiting, vital2024predicting}. Initial methods used handcrafted features, such as author reputation, venue rank, early citations, with regression or classification models, but struggled to capture the complexity of impact dynamics \citep{ibanez2009predicting}. More recent approaches adopt data-driven and network-aware models, including temporal models that simulate citation growth via paper ``fitness'' and decay \citep{wang2013quantifying}. \cite{jiang2021hints} proposed HINTS, an end-to-end model predicting citation time series from publication time using pre-publication metadata and bibliographic networks, effectively addressing the cold-start problem and outperforming models dependent on years of citation data. \cite{xue2024predicting} introduced a GNN-based framework leveraging dynamic citation graphs and auxiliary tasks to yield interpretable and accurate predictions across paper lifespans and disciplines. These developments highlight the value of modeling scientific impact as a multifactorial process rather than a singular metric.

\paragraph{Heterogeneous Scholarly Networks} Scholarly impact features are effectively modeled as heterogeneous information networks (HIN), where nodes (e.g., papers, authors, venues) and edges (e.g., citations, co-authorship) represent diverse entities and relations  \citep{geng2022modeling, he_h2cgl_2023}. Meta-path analysis enabled early relation-specific influence measures \citep{sun2011pathsim}, followed by ranking methods that integrated content, venue, and publication networks \citep{tang2008arnetminer}. More recent work uses temporal GNNs and embedding alignment to track evolving influence and predict impact by embedding papers into historical network contexts \citep{holm_longitudinal_2021}. Advanced models apply relational GNNs with attention and temporal encoders to weigh author, venue, and content signals differently over time. HIN-based methods have also addressed collaboration prediction, topic emergence, and prestige estimation, confirming their strength for multi-scale scientific impact analysis \citep{zhao2025words, de2024can, arts_beyond_2024, jin_scito2m_2024}. These models capture network structure and temporal context but underutilize the semantic richness of texts.

\paragraph{LLMs for Scientific Understanding.} With the fast development of LLMs, researchers are increasingly exploring their ability to predict scientific impact from textual content alone \citep{cohan2020specter, de2024can, zhao2025words}. Early efforts used simple representations like TF-IDF, but newer models leverage transformers trained on scientific corpora. SciBERT \citep{beltagy-etal-2019-scibert} improved classification and recommendation by capturing domain-specific language, while SPECTER \citep{cohan2020specter} showed that content-based embeddings correlate with scholarly relevance. \cite{zhao2025words} proposed a content-only LLM framework predicting impact from titles and abstracts in a double-blind fashion, achieving state-of-the-art performance using a field- and time-normalized metric, TNCSI$_{SP}$. Similarly, \cite{vital2024predicting} showed GPT-based abstract embeddings could effectively identify highly cited papers, and even TF-IDF performed competitively, underscoring the role of topical relevance. \cite{de2024can} found ChatGPT-4’s qualitative scores on novelty, clarity, and engagement significantly correlated with later citations and Altmetric scores, suggesting LLMs can evaluate intangible manuscript qualities linked to impact. These studies show that LLMs excel at capturing intrinsic semantic merit and can complement or even rival traditional metadata-driven models. Combining LLM-derived content features with graph-based signals promises a holistic, multi-scale approach to modeling scientific influence.

\section{Dataset Construction}
\label{sec:dataset_construction}

To support systematic research on impact prediction, we construct a large-scale benchmark from the Semantic Scholar database ~\citep{Kinney2023TheSS}, along with an automatic pipeline for its construction. 

\subsection{Construction Pipeline}
\label{sec:construction_pipeline}

\paragraph{Source corpus and query sampling.}

We build on the full S2AG snapshot, comprising 237{,}026{,}483 papers published between 1970 and 2025. Each paper is associated with static metadata, including publication information, venue, citation statistics, field-of-study tags, and abstract text. We define a \emph{query paper} as a paper published between 2005 and 2020 with a non-empty reference list and an available abstract, ensuring both graph structure and textual content for downstream prediction. From the eligible pool, we uniformly sample 2{,}000{,}000 query papers without field-based stratification. We then split them temporally into a \emph{historical} bucket (2005--2017) and a \emph{fresh} bucket (2018--2020), with \texttt{train}, \texttt{val}, and \texttt{previous-test} drawn from the former and \texttt{fresh-test} from the latter. The temporal split is performed after the 2M papers are sampled, so the resulting historical and fresh subsets follow the natural temporal distribution of the sampled corpus. This temporal split enables evaluation of generalization to more recent papers and their citations.

\begin{wraptable}{r}{0.45\textwidth}
\vspace{-5mm}
\centering
\small
\caption{\textbf{Statistics of our benchmark}: 2M papers with temporally separated test sets.}
\label{tab:dataset_stats}
\resizebox{0.44\textwidth}{!}{
\begin{tabular}{lrr}
\toprule
\textbf{Split} & \textbf{Queries} & \textbf{Month Zero labels} \\
\midrule
Train          & 800K & 722{,}510 (90.3\%) \\
Validation     & 300K & 270{,}810 (90.3\%) \\
Previous Test  & 400K & 361{,}196 (90.3\%) \\
Fresh Test     & 500K & 467{,}835 (93.6\%) \\
\midrule
\multicolumn{3}{l}{Query papers: 2M \quad Years: 2005--2020} \\
\multicolumn{3}{l}{Ego-graph: 39.3M nodes, 1.18B edges} \\
\multicolumn{3}{l}{Metadata: author 99.5\%, venue 46.1\%, ROR 7.7\%} \\
\bottomrule
\end{tabular}}
\vspace{-5mm}
\end{wraptable}

\paragraph{Ego-graph construction and leakage-safe features.}

For each paper, we extract its 1-hop references and expand each 1-hop node with its top-25 references by citation count, forming a 2-hop ego-graph. To prevent temporal leakage, all citation counts are truncated to their values as of the query paper's publication year. Ego-graph statistics are reported in Table~\ref{tab:dataset_stats}.



\paragraph{Metadata cleaning and quality control.}
We augment each paper with author identities, canonical venues, and institutions, using S2AG's native venue links and ROR-based affiliation matching. We validate the resulting dataset through structural and referential consistency checks, including split disjointness, label monotonicity, citation-count validity, and key uniqueness. 
To rule out \textbf{memorization leakage}, all LLMs are queried without web search, and zero-shot LLMs perform worst among all methods (Tables~\ref{tab:main_previous}--\ref{tab:main_fresh}), indicating that they cannot recall citation counts.

\subsection{Dataset Statistics}
\label{sec:dataset_statistics}

\paragraph{Prediction targets.}
Following common formulations of citation-based impact prediction---single-horizon forecasting \citep{jiang2021hints}, multi-step citation trajectories \citep{xue2024predicting}, and newborn-paper prediction \citep{zhao2025words}---we consider both year-level ($k=1,\ldots,5$) and month-level ($k=1,\ldots,5$) citation counts as prediction targets. We provide both per-period and \emph{cumulative} citation counts. For papers without a complete \texttt{publicationdate}, month-level targets are marked as \texttt{None} rather than zero, since the absence of a valid date does not imply zero citations.

\begin{wrapfigure}{r}{0.5\textwidth}
\vspace{-\intextsep}
  \centering
  \begin{subfigure}{0.49\linewidth}
    \centering
    \includegraphics[width=\linewidth]{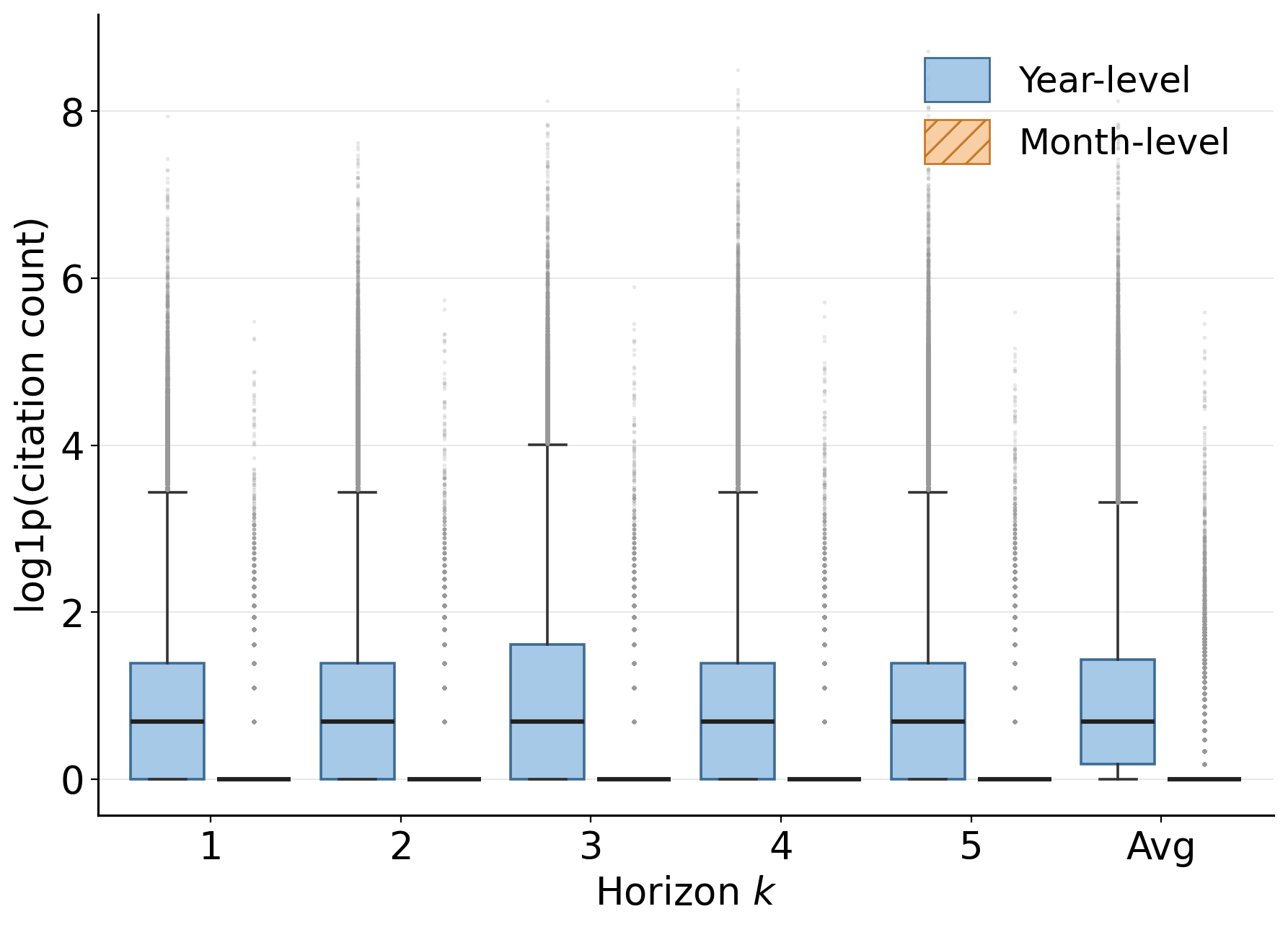}
    \caption{Citation Distribution}
    \label{fig:year_month_boxplot}
  \end{subfigure}
  \hfill
  \begin{subfigure}{0.49\linewidth}
    \centering
    \includegraphics[width=\linewidth]{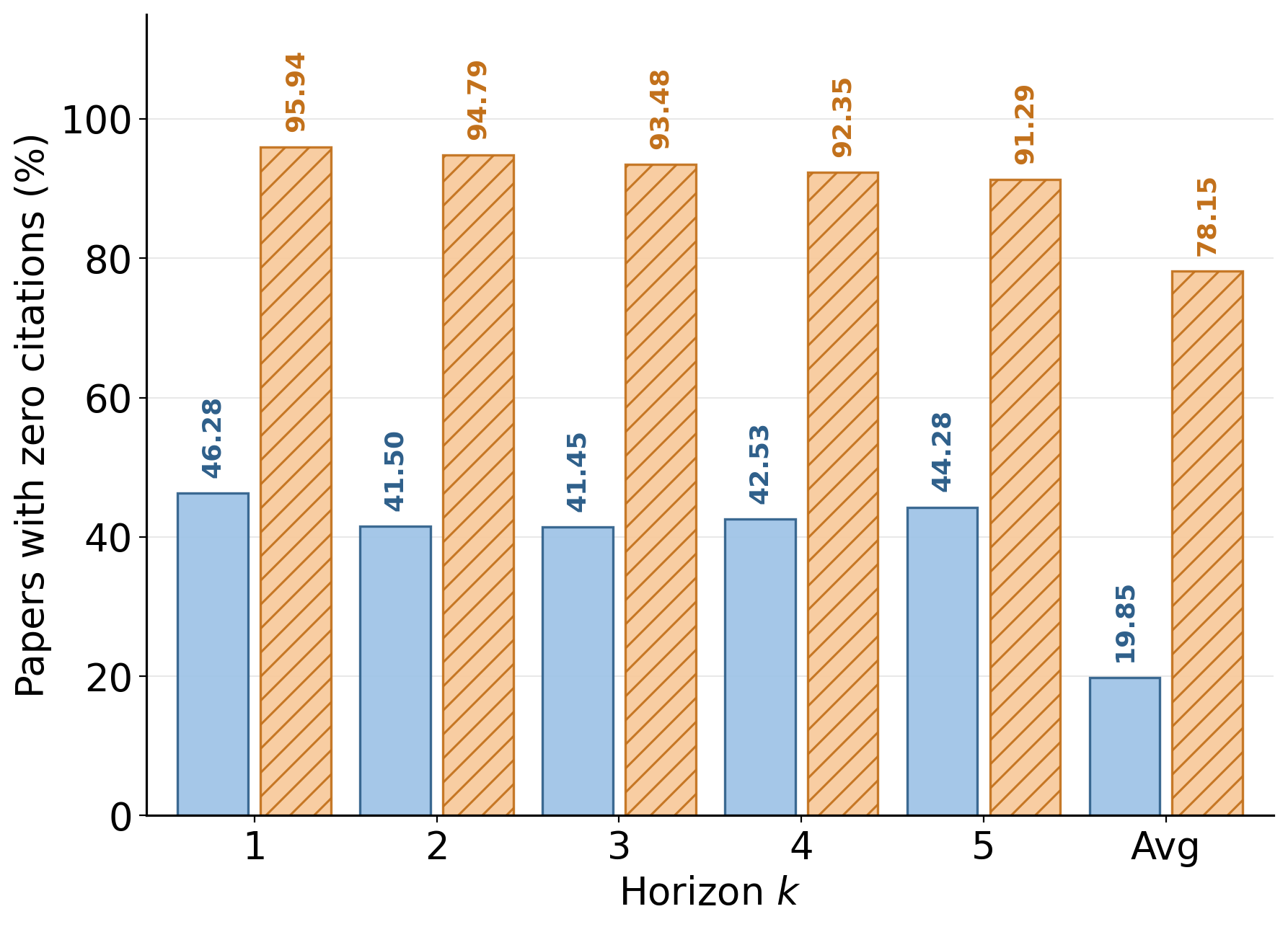}
    \caption{Zeros citation ratio}
    \label{fig:ym_zero}
  \end{subfigure}
  \caption{\textbf{Statistics of citation numbers.}}
  \label{fig:year_month}
\vspace{-\intextsep}
\end{wrapfigure}

\paragraph{Month-level targets are highly zero-inflated.}
\revision{Year- and month-level targets exhibit substantially different distributions (Figure~\ref{fig:year_month}). 
Unlike year-level counts, month-level targets remain zero for most papers at every horizon. Under such zero inflation, near-zero predictions achieve low regression error without discriminating among cited papers, so we complement regression metrics with threshold-sensitive classification metrics for the month-level task.}
\section{Method}
\label{sec:method}

\begin{figure}[t]
    \centering
    \includegraphics[width=0.96\textwidth]{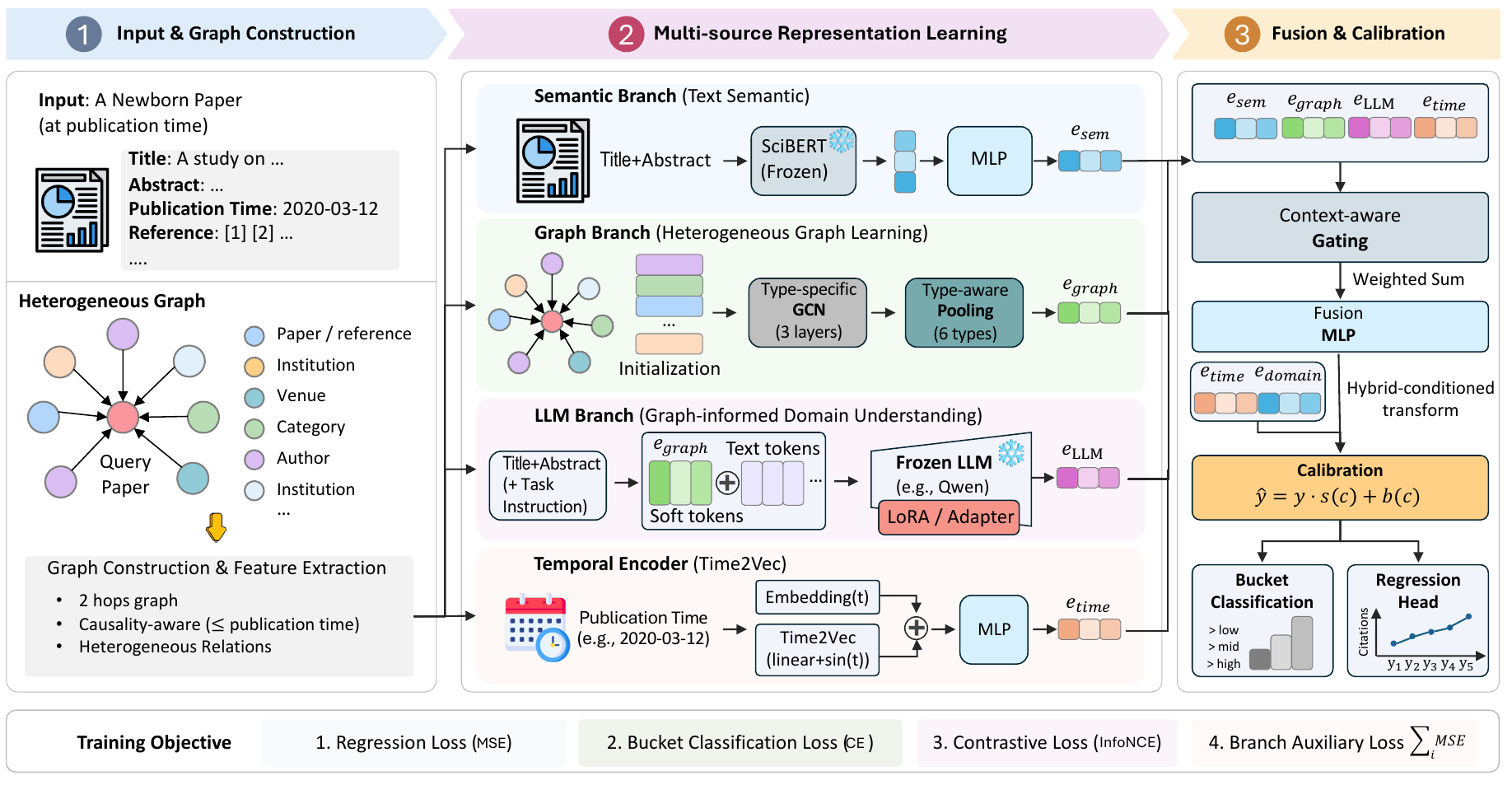}
    \caption{\textbf{Our proposed LLM4Impact.} A cold-start impact prediction method that integrates semantic, heterogeneous graph, and temporal evidence. Graph structure is kept as a separate branch and also injected into a frozen LLM via continuous prefix tokens, followed by domain- and time-aware gated fusion and decoupled context-conditioned calibration.}
    \label{fig:framework_overview}
    \vspace{-5mm}
\end{figure}

Scientific impact prediction is a cold-start problem: a newly published paper has no citation history and must rely on structural, semantic, and temporal evidence. 
Thus, we propose \emph{LLM4Impact} (Figure~\ref{fig:framework_overview}), which integrates these signals through multi-branch and context-aware fusion.

Our design addresses \textbf{three distinct challenges} in cold-start impact prediction: representing unseen publication years, adapting to instance-dependent evidence reliability, and correcting domain- and time-dependent citation scales. We address them with \textbf{three corresponding components}: (i) a diagnostic-driven temporal encoder with continuous extrapolation and a zero-initialized residual, (ii) domain- and time-aware gated fusion, and (iii) a decoupled context-conditioned calibration module.

\subsection{Task Formulation}
\label{sec:task}

We formulate scientific impact prediction as a cold-start regression problem over a heterogeneous graph. Let $G=\langle N,E\rangle$ denote the heterogeneous citation graph, where nodes represent papers, authors, institutions, venues, categories, references, and publication time, and edges encode authorship, affiliation, venue, citation, and publication-time relations.

For a query paper $p$, all graph information is restricted to what was available at its publication time: its own citation count is zero, and neighboring papers are assigned only their point-in-time citation counts. This prevents future information from leaking into the prediction. Given $p$, the goal is to predict its cumulative citation counts over $L$ future horizons:

\begin{equation}
Y_p = \{y_p^1,\ldots,y_p^L\},
\label{eq:task}
\end{equation}
where $y_p^\ell$ denotes the cumulative citation count at horizon $\ell$. We consider yearly prediction ($L=5$) and monthly prediction ($L=5$, month 1 through month 5). The model jointly leverages the query paper's heterogeneous graph, semantic content, and publication time.

\subsection{Multi-source Representation Learning}

Each query paper is represented by three complementary branches: semantic, graph, and temporal. Each branch produces a $d$-dimensional representation, with fusion deferred until Section~\ref{sec:fusion}.

\paragraph{Semantic branch.}
We encode the title and abstract of $p$ using a frozen SciBERT~\citep{beltagy-etal-2019-scibert} encoder followed by a two-layer projection head. The resulting representation is used as the semantic embedding of the query paper and also initializes paper/reference nodes in the graph branch, grounding semantic and structural representations in the same feature space.

\paragraph{Heterogeneous graph branch.}
We encode the query paper's 2-hop heterogeneous ego-graph using type-specific feature projections followed by a citation-aware GNN. Citation edges are weighted according to the sender's point-in-time citation count, while other relation types use mean aggregation. Type-aware attention pooling summarizes the heterogeneous neighborhood into a compact graph representation. During training, we additionally apply contrastive learning \citep{oord2019representationlearningcontrastivepredictive} between clean and popularity-augmented graph views. 

\subsection{Graph-to-LLM Prefix Injection}
\label{sec:llm}

The graph representation provides structural evidence that is difficult to express through the LLM's text input alone. We therefore project the graph representation into the LLM hidden space and reshape it into $L_p$ continuous prefix tokens \citep{li-liang-2021-prefix}:
\begin{equation}
\mathbf{P}
=
\operatorname{reshape}
\left(
\mathcal{P}_{\mathrm{graph}}
(\mathbf{e}_{\mathrm{graph}})
\right)
\in
\mathbb{R}^{L_p\times d_{\mathrm{LLM}}}.
\label{eq:prefix}
\end{equation}

The prefix is prepended to the tokenized task instruction, title, and abstract directly in embedding space, while the LLM backbone remains frozen. The text positions are then pooled to obtain the LLM representation. Only the prefix projector, lightweight adaptation layers, and projection are trainable.

Prefix injection itself has been explored in other graph-centric tasks, but its effectiveness for citation-impact prediction has not been systematically established. The distinction is important because the query paper is cold-start and has no citation history of its own. We therefore compare embedding-space prefix injection against natural-language serialization of the same structural information. The prefix formulation consistently outperforms textual injection, supporting continuous structural conditioning as an effective interface for this setting.

\subsection{Diagnosing and Fixing Temporal Encoding}
\label{sec:temporal}

Publication time is essential for impact prediction, but a standard lookup-based time embedding can fail under temporal generalization. Our fresh test covers papers from 2018--2020, while training uses 2005--2017. Thus, a year-based lookup assigns an untrained embedding to every fresh-test paper, since these years receive no gradient updates during training. We verify that this is the primary source of the fresh-test degradation rather than graph sparsity. Specifically, the error--graph-density relationship is nearly identical across the historical and fresh splits, ruling out graph density as the main differentiating factor and pointing instead to the unseen publication-time representation.

\revision{We therefore replace the lookup-only representation with a continuous, extrapolable time encoder \citep{kazemi2019time2veclearningvectorrepresentation} that applies uniformly to both year- and month-level prediction. Let $\tau_p$ denote the publication time of paper $p$, measured at the corresponding temporal granularity (year or month), and let $t_p$ be its normalized value. We encode publication time as}
\begin{equation}
\mathbf{e}_{\mathrm{time}}
=
\mathrm{MLP}_{\mathrm{time}}
\left(
[
\sin(2\pi t_p\boldsymbol{\omega});
\cos(2\pi t_p\boldsymbol{\omega})
]
\right)
+
\mathrm{Embed.}_{\mathrm{time}}[\tau_p]
\qquad
t_p =
\frac{\tau_p-\tau_{\min}}
     {\tau_{\max}-\tau_{\min}}.
\label{eq:temporal}
\end{equation}

This provides an extrapolable temporal representation, while the zero-initialized residual captures time-specific deviations for observed timestamps. Unseen timestamps simply fall back to the continuous base representation, enabling generalization across year- and month-level prediction.

\subsection{Domain- and Time-Aware Gated Fusion}
\label{sec:fusion}

Evidence sources vary in their informativeness across papers. Thus, we learn an instance-specific gate conditioned on publication time and research domain. For paper $p$, we define $\mathbf{c}=[\mathbf{e}_{\mathrm{time}}\Vert\mathbf{d}_p]$, where $\mathbf{d}_p$ is the mean-pooled category embedding. The gate produces normalized weights over branches:

\begin{equation}
\mathbf{g}
=
\operatorname{softmax}
\left(
W_g
[
\mathbf{e}_1\Vert\cdots\Vert\mathbf{e}_n
\Vert\mathbf{c}
]
\right).
\label{eq:gating}
\end{equation}

The gated branch mixture is then combined with the joint branch representation through a fusion MLP. Unlike fixed-weight fusion, the learned gate allows the model to adapt which evidence source to trust according to the paper's domain and publication era.

\subsection{Decoupled Prediction Calibration}
\label{sec:calibration}

\revision{Evidence selection and prediction calibration address different problems. While fusion determines which evidence sources to trust, citation scales vary systematically across research domains and publication cohorts. We therefore apply a separate context-conditioned calibration module:}
\begin{equation}
\hat{\mathbf{y}}
=
\hat{\mathbf{y}}_{\mathrm{fused}}
\odot
\left(1+\operatorname{MLP}_{s}(\mathbf{c})\right)
+
\operatorname{MLP}_{b}(\mathbf{c}).
\label{eq:calibration}
\end{equation}

Zero-initializing the final calibration layers makes the module an identity transformation at initialization. Fusion therefore learns \emph{which evidence to trust}, while calibration learns \emph{how to rescale the estimate} based on domain and publication time. 
We train the model end-to-end with a composite objective:

\begin{equation}
\mathcal{L}
=
\mathcal{L}_{\mathrm{reg}}
+
\lambda_{\mathrm{aux}}\mathcal{L}_{\mathrm{bucket}}
+
\lambda_{\mathrm{cl}}\mathcal{L}_{\mathrm{contrast}}
+
\lambda_{\mathrm{branch}}\mathcal{L}_{\mathrm{branch}}.
\label{eq:objective}
\end{equation}

The main regression loss operates in $\log(1+x)$ space. The auxiliary terms provide coarse citation-bucket supervision, graph-level contrastive learning, and branch-wise deep supervision, respectively. Detailed loss definitions and training settings are provided in Appendix~\ref{ax:method_details}.
\section{Experiment and Analysis}

\subsection{Experiment Setting}

\paragraph{Baselines} 

We compare against four representative baselines: SciBERT, a scientific text encoder; HINTS~\citep{jiang2021hints}, a dynamic heterogeneous network model; and NAIP~\citep{zhao2025words}, an LLM-based impact
prediction method. We additionally evaluate Qwen2.5-7B \citep{qwen2025qwen25technicalreport}, Qwen3-0.6B \citep{yang2025qwen3}, Qwen3.5-4B, and Qwen3.5-9B~\citep{qwen3.5} via prompting and fine-tuning, excluding models larger than 9B due to dataset scale. Finally, we adapt H2CGL~\citep{he_h2cgl_2023} and DPPDCC~\citep{xue2024predicting} to the newborn-paper setting to avoid citation leakage.

\paragraph{Evaluation Metrics} 
We evaluate performance using regression, ranking, and zero-inflation-aware metrics. Year-level prediction is evaluated with RMSE. For month-level prediction, where 90.3\% of \texttt{previous\_test} labels are zero (see Table~\ref{tab:dataset_stats} and Section~\ref{sec:dataset_statistics}), we additionally report AUROC, PR-AUC, and QWK \citep{Cohen1968}, together with RMSE and NDCG@20. RMSE measures numerical prediction error, while NDCG@20 evaluates the quality of the top-20 ranked citation lists. Higher is better for AUROC, PR-AUC, QWK, and NDCG@20; lower is better for RMSE. Detailed definition of these metrics are in Appendix~\ref{ax:metric_definition}.

\subsection{Main Results}

Tables~\ref{tab:main_previous} and~\ref{tab:main_fresh} summarize the performance of our proposed LLM4Impact on the in-distribution and temporally shifted test sets, respectively. Overall, \textbf{LLM4Impact consistently outperforms existing approaches across both year- and month-level prediction, demonstrating the benefit of jointly modeling semantic, graph, and temporal evidence.}

On \emph{previous\_test}, our 7B model achieves the best performance at all five year-level horizons, reducing average RMSE from 0.790 of the strongest prior baseline, NAIP, to 0.710 (\textbf{10.13\%} improvement). The gain grows with the prediction horizon, from 7.38\% at $Y_1$ to 11.40\% at $Y_5$, suggesting that structured and temporal evidence becomes increasingly valuable for longer-term impact prediction. \revision{The gains also extend to the highly zero-inflated month-level setting, where our model achieves the best trained-model performance across AUROC, PR-AUC, QWK, and RMSE, with a particularly large improvement in QWK (+40.77\%)}. The smaller Qwen3-0.6B variant also outperforms prior baselines on most metrics, indicating that the gains are not solely attributable to the larger backbone.

More importantly, LLM4Impact maintains its advantage under temporal distribution shift. On \emph{fresh\_test}, where publication years are unseen during training, our 7B model remains the best trained approach at every year-level horizon, achieving an average RMSE of 0.786 and a 6.87\% improvement over NAIP. It also achieves the best month-level performance across the reported metrics, including an 11.73\% improvement in NDCG. This consistent performance on unseen publication years supports the effectiveness of our continuous temporal encoding and context-conditioned calibration. In contrast, zero-shot LLMs generally perform substantially worse on year-level prediction, highlighting that general-purpose language modeling alone is insufficient for cold-start scientific impact prediction.

\begin{table*}[t]
\centering
\caption{\textbf{Main results on previous-test (in-distribution).} Year-level results use per-horizon RMSE$\downarrow$ in log1p space, while month-level results use zero-inflation-aware metrics due to 90.3\% zero labels. Zero-shot LLMs are reported separately. Best results among trained models are shown in \textbf{bold}.}
\label{tab:main_previous}
\resizebox{\textwidth}{!}{%
\begin{tabular}{l cccccc ccccc}
\toprule
\multirow{2}{*}{\textbf{Method}} 
& \multicolumn{6}{c}{\textbf{Year} (RMSE $\downarrow$)} & \multicolumn{5}{c}{\textbf{Month}} \\
\cmidrule(lr){2-7} \cmidrule(lr){8-12}
 & Y1 & Y2 & Y3 & Y4 & Y5 & Avg & AUROC$\uparrow$ & PR-AUC$\uparrow$ & QWK$\uparrow$ & RMSE$\downarrow$ & NDCG$\uparrow$ \\
\midrule
\multicolumn{12}{l}{\textit{Traditional}} \\
\midrule
SciBERT \citep{beltagy-etal-2019-scibert}  & 0.647 & 0.779 & 0.858 & 0.916 & 0.958 & 0.839 & 0.783 & 0.406 & 0.095 & 0.296 & 0.562 \\
HINTS~\citep{jiang2021hints}     & 0.709 & 0.857 & 0.945 & 1.008 & 1.055 & 0.923 & 0.621 & 0.196 & 0.000 & 0.323 & 0.000 \\
H2CGL \citep{he_h2cgl_2023}    & 0.626 & 0.741 & 0.813 & 0.864 & 0.905 & 0.796 & 0.781 & 0.417 & 0.130 & 0.296 & 0.246 \\
DPPDCC \citep{xue2024predicting}   & 0.665 & 0.791 & 0.866 & 0.920 & 0.958 & 0.846 & 0.752 & 0.364 & 0.036 & 0.305 & 0.383 \\
\midrule
\multicolumn{12}{l}{\textit{*LLMs-based*}} \\
\midrule
NAIP (7B)~\citep{zhao2025words} & 0.623 & 0.738 & 0.806 & 0.857 & 0.895 & 0.790 & 0.765 & 0.397 & 0.114 & 0.299 & 0.118 \\
Qwen3-0.6B (zs) \citep{yang2025qwen3}  & 1.146 & 1.629 & 1.942 & 2.164 & 2.329 & 1.889 & 0.500 & 0.148 & 0.001 & 0.360 & 0.000 \\
Qwen3.5-4B (zs) \citep{qwen3.5}  & 1.189 & 1.494 & 1.694 & 1.820 & 1.912 & 1.642 & 0.648 & 0.249 & 0.099 & 1.245 & 0.090 \\
Qwen3.5-9B (zs) \citep{qwen3.5} & 1.670 & 1.966 & 2.108 & 2.192 & 2.252 & 2.048 & 0.678 & 0.257 & 0.045 & 1.746 & 0.432 \\
\midrule
Ours (Qwen3-0.6B) & 0.612 & 0.719 & 0.782 & 0.829 & 0.860 & 0.766 & 0.797 & 0.439 & 0.144 & 0.290 & 0.436 \\
\textbf{Ours}(Qwen2.5-7B) & \textbf{0.577} & \textbf{0.670} & \textbf{0.725} & \textbf{0.763} & \textbf{0.793} & \textbf{0.710} & \textbf{0.802} & \textbf{0.450} & \textbf{0.183} & \textbf{0.286} & \textbf{0.566} \\
\rowcolor{gray!15}
\textit{Improvement} & \textit{+7.38\%} & \textit{+9.21\%} & \textit{+10.05\%} & \textit{+10.97\%} & \textit{+11.40\%} & \textit{+10.13\%} & \textit{+2.43\%} & \textit{+7.91\%} & \textit{+40.77\%} & \textit{+3.38\%} & \textit{+0.71\%} \\
\bottomrule
\end{tabular}%
}
\end{table*}

\begin{table*}[t]
\centering
\caption{\textbf{Main results on fresh-test (out-of-distribution, temporally-shifted papers).} Same column design as Table~\ref{tab:main_previous}. Best result per column among trained models in \textbf{bold}.}
\label{tab:main_fresh}
\resizebox{\textwidth}{!}{%
\begin{tabular}{l cccccc ccccc}
\toprule
& \multicolumn{6}{c}{\textbf{Year} (RMSE $\downarrow$)} & \multicolumn{5}{c}{\textbf{Month}} \\
\cmidrule(lr){2-7} \cmidrule(lr){8-12}
Model & Y1 & Y2 & Y3 & Y4 & Y5 & Avg & AUROC$\uparrow$ & PR-AUC$\uparrow$ & QWK$\uparrow$ & RMSE$\downarrow$ & NDCG$\uparrow$ \\
\midrule
\multicolumn{12}{l}{\textit{Traditional}} \\
\midrule
SciBERT   \citep{beltagy-etal-2019-scibert} & 0.728 & 0.848 & 0.919 & 0.977 & 1.019 & 0.904 & 0.747 & 0.399 & 0.049 & 0.355 & 0.588 \\
HINTS  ~\citep{jiang2021hints}    & 0.773 & 0.902 & 0.970 & 1.018 & 1.054 & 0.949 & 0.633 & 0.246 & 0.000 & 0.376 & 0.015 \\ 
H2CGL  \citep{he_h2cgl_2023}     & 0.713 & 0.838 & 0.922 & 0.975 & 1.016 & 0.899 & 0.762 & 0.422 & 0.090 & 0.351 & 0.259 \\
DPPDCC  \citep{xue2024predicting}   & 0.743 & 0.869 & 0.940 & 0.989 & 1.037 & 0.921 & 0.697 & 0.346 & 0.023 & 0.367 & 0.229 \\
\midrule
\multicolumn{12}{l}{\textit{*LLMs-based*}} \\
\midrule
NAIP (7B)~\citep{zhao2025words} & 0.698 & 0.802 & 0.862 & 0.904 & 0.935 & 0.844 & 0.729 & 0.389 & 0.053 & 0.356 & 0.298 \\
Qwen3-0.6B (zs) \citep{yang2025qwen3}  & 1.262 & 1.736 & 2.016 & 2.211 & 2.360 & 1.956 & 0.500 & 0.183 & 0.001 & 0.420 & 0.000 \\
Qwen3.5-4B (zs) \citep{qwen3.5}  & 1.268 & 1.591 & 1.794 & 1.916 & 2.001 & 1.734 & 0.651 & 0.298 & 0.112 & 1.313 & 0.039 \\
Qwen3.5-9B (zs) \citep{qwen3.5}  & 1.661 & 1.962 & 2.126 & 2.228 & 2.296 & 2.067 & 0.680 & 0.302 & 0.053 & 1.766 & 0.074 \\ 
\midrule
Ours (Qwen3-0.6B) & 0.699 & 0.801 & 0.851 & 0.889 & 0.915 & 0.835 & 0.759 & 0.429 & 0.078 & 0.352 & 0.524 \\
\textbf{Ours} (Qwen2.5-7B) & \textbf{0.657} & \textbf{0.748} & \textbf{0.801} & \textbf{0.839} & \textbf{0.870} & \textbf{0.786} & \textbf{0.771} & \textbf{0.441} & \textbf{0.112} & \textbf{0.347} & \textbf{0.657} \\
\rowcolor{gray!15}
\textit{Improvement} & \textit{+5.87\%} & \textit{+6.73\%} & \textit{+7.08\%} & \textit{+7.19\%} & \textit{+6.95\%} & \textit{+6.87\%} & \textit{+1.18\%} & \textit{+4.50\%} & \textit{+24.44\%} & \textit{+1.14\%} & \textit{+11.73\%} \\

\bottomrule
\end{tabular}%
\vspace{-5mm}
}
\end{table*}

\begin{wraptable}{l}{0.55\columnwidth}
\centering
\caption{\textbf{Ablation at year granularity.} $\Delta$ denotes the relative change compared with the full model.}
\label{tab:ablation}

\resizebox{\linewidth}{!}{%
\begin{tabular}{llccccc cc}
\toprule
Split & Model & Y1 & Y2 & Y3 & Y4 & Y5 & Avg. RMSE & $\Delta$ \\
\midrule
\multirow{4}{*}{\emph{previous\_test}}
& \textbf{Ours (7B)}
& \textbf{0.577} & \textbf{0.670} & \textbf{0.725} & \textbf{0.763} & \textbf{0.793}
& \textbf{0.710} & -- \\
& $-$ Calibration
& 0.618 & 0.729 & 0.794 & 0.843 & 0.878
& 0.778  & +9.6\% \\
& $-$ LLM branch
& 0.628 & 0.746 & 0.817 & 0.868 & 0.908
& 0.799 & +12.6\% \\
& $-$ Graph branch
& 0.602 & 0.711 & 0.774 & 0.819 & 0.853
& 0.757 & +6.6\% \\
\midrule
\multirow{4}{*}{\emph{fresh\_test}}
& \textbf{Ours (7B)}
& \textbf{0.657} & \textbf{0.748} & \textbf{0.801} & \textbf{0.839} & \textbf{0.870}
& \textbf{0.786} & -- \\
& $-$ Calibration
& 0.702 & 0.814 & 0.881 & 0.937 & 0.976
& 0.868 & +10.4\% \\
& $-$ LLM branch
& 0.708 & 0.829 & 0.905 & 0.962 & 1.005
& 0.888 & +12.9\% \\
& $-$ Graph branch
& 0.679 & 0.777 & 0.835 & 0.878 & 0.908
& 0.819 & +4.2\% \\
\bottomrule
\end{tabular}%
}
\end{wraptable}

\subsection{Ablation Study}

\paragraph{Component Effectiveness.}
Table~\ref{tab:ablation} evaluates the contribution of the major components on both in-distribution and temporally shifted data. Re-training method without LLM branch causes the largest degradation, increasing average RMSE by 12.6\% on \emph{previous\_test} and 12.9\% on \emph{fresh\_test}, demonstrating the value of LLM-based semantic evidence. Re-training method without calibration increases RMSE by 9.6\% and 10.4\%, respectively, supporting the benefit of decoupling prediction adjustment from evidence fusion. Re-training method without graph branch results in consistent increases of 6.6\% and 4.2\%, confirming the complementary role of structural evidence. The consistent effects across the two splits indicate that these components provide complementary benefits that persist under temporal distribution shift.

\paragraph{Sensitivity to Perturbations.}

\begin{figure}[t]
    \centering
    \includegraphics[width=0.99\textwidth]{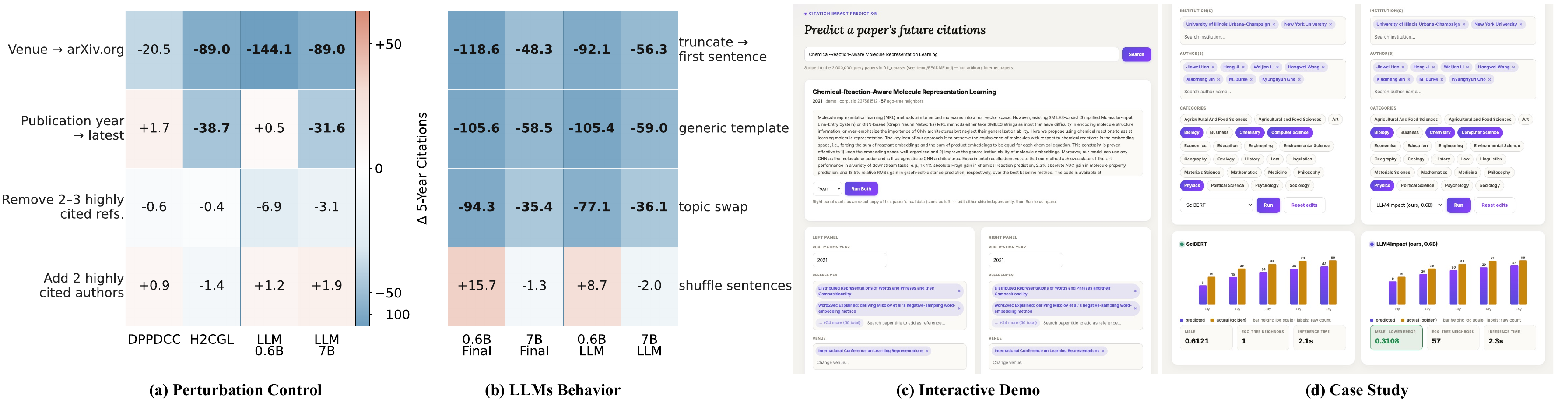}
    \caption{\textbf{Robustness, Interpretability, and Practical Validation of LLM4Impact.} (a) Contribution of different evidence sources; (b) LLM behavior under controlled perturbations; (c) Interactive exploration of citation predictions; (d) Case study comparing LLM4Impact with SciBERT.}
    \label{fig:ablation_analysis}
    \vspace{-5mm}
\end{figure}

\revision{To examine what drives citation predictions and validate our multi-source framework, we conduct two complementary controlled perturbation studies on the same 20 computer science papers. As shown in Figure~\ref{fig:ablation_analysis}, (a) probes metadata sensitivity by perturbing authors, publication year, venue, and highly cited references while keeping other information fixed. Venue changes produce the largest effects across models, while author perturbations have much smaller impacts. The varying responses to publication year and citation structure further motivate domain- and time-aware gated fusion. (b) probes the LLM branch through sentence shuffling, truncation, generic replacement, and topic swapping. Content removal or replacement substantially changes both isolated LLM and final fused predictions, whereas sentence shuffling has a much smaller effect. These results indicate that the LLM branch captures content-level signals and provides complementary evidence to the graph and semantic branches, supporting our multi-source fusion design.}

\subsection{More Analysis and discussion}
\label{sec:more_analysis}

\paragraph{Case Study: Interactive Sensitivity and Counterfactual Analysis}


We further provide an interactive demo for probing how contextual factors
affect predicted scientific impact. Users can independently modify
publication year, references, venue, authors, institutions, and research
categories, while keeping the remaining information fixed, and compare the
resulting citation trajectories across models. This enables controlled
what-if analysis of how changes in individual factors alter the model's
predicted impact, going beyond measuring sensitivity to input perturbations.
The supplementary webpage provides additional screenshots and more case studies
are in Appendix~\ref{ax:more_case_study}.

\begin{wraptable}{r}{0.5\linewidth}
\small
\caption{\textbf{Comparison of RMSE and correlation \textit{Cor.}} between C\_log and C\_TNCSI on \textit{fresh} set.}
\label{tb:comparsion_of_tncsi_fresh}
\resizebox{0.5\textwidth}{!}{
\begin{tabular}{l|ccccc|c}
\toprule
Metric   & Year\_1      & Year\_2      & Year\_3      & Year\_4      & Year\_5      & Avg    \\ \midrule
C\_log   & {0.7170} & {0.7426} & {0.7892} & {0.8302} & {0.9514} & {0.7745}  \\ \midrule
C\_TNCSI & 0.2316 & 0.2441 & 0.2556 & 0.2556 & 0.2532 & 0.2480 \\ \midrule
\textbf{Cor.}     &  \textbf{0.8004} & \textbf{0.8128} & \textbf{0.8095} & \textbf{0.8136} & \textbf{0.8128} & \textbf{0.8098} \\ 
\bottomrule
\end{tabular}}
\end{wraptable}

\paragraph{Other Metrics.} We further investigated the impact of alternative metrics on model evaluation. Following prior work, we extended our prediction target to Topic Normalized Citation Success Index(TNCSI) \citep{zhao2025words}, which normalizes citation counts relative to the citation distribution of publications in the same field, thus mitigating cross-disciplinary citation bias. Table \ref{tb:comparsion_of_tncsi_fresh} presents the RMSE scores via using both raw citation counts and the TNCSI metric, and their Spearman correlation coefficients \citep{spearman1961proof}. We observe that despite differences in relative scores, the corresponding predictions exhibit a \textbf{strong positive correlation}. Besides, we find that the fresh test set continues to pose greater challenges compared to the previous one.
\section{Conclusion}

In this paper, we presented \emph{LLM4Impact}, a unified framework for cold start scientific impact prediction that integrates semantic, heterogeneous graph, LLM, and temporal evidence. By combining continuous temporal representation, context aware evidence gating, and context conditioned calibration, LLM4Impact consistently improves over strong baselines across yearly and monthly prediction tasks. Ablation and perturbation studies further show that different evidence sources provide complementary information and remain useful across changing publication contexts. 


\section*{Acknowledgments}
Special thanks to the AVG members for their help and discussions. Prof. Andreas Geiger are members of the Machine Learning Cluster of Excellence, EXC number 2064/1 – Project number 390727645. Andreas Geiger was supported by the ERC Starting Grant LEGO-3D (850533). Yong Cao was supported by a VolkswagenStiftung Momentum grant.

\section*{AI use statement}
Large language models are used in our early-stage writing for wording and grammar checking, as well as searching for missing literature. LLMs are not involved in the later iterations of the paper writing. Therefore, we do not consider LLMs as significant contributors to this paper.

\subsubsection*{Reproducibility statement}
We will publicly release our full code, including training, inference, and evaluation scripts, to ensure that all results in this paper are fully reproducible. 

\bibliography{references}
\bibliographystyle{iclr2027_conference}

\clearpage
\appendix

\makeatletter
\renewcommand{\partname}{}
\renewcommand{\thepart}{}
\makeatother

\part{Appendix} 
{
  \hypersetup{linkcolor=black}
  \parttoc
}

\section{Dataset Construction and Statistics}
\label{ax:dataset_details}

\begin{wraptable}{r}{0.48\linewidth}
\centering
\caption{Comparison of popular impact prediction paper datasets and databases.
Abbreviations: Gran. = multi-grained (year- and month-level) labels, Dom. = multi-domain coverage.
$^\dagger$Our domain coverage is the \emph{natural}, unstratified distribution of eligible S2AG papers, not a balanced sample -- see Table~\ref{tab:dataset_stats} for the resulting shift relative to the full corpus.}
\label{tab:dataset_comparison}
\resizebox{0.48\textwidth}{!}{
\begin{tabular}{lcccc c}
\toprule
\textbf{Dataset} & Year & Gran. & Dom. & Tools & \#Papers \\
\midrule
APS       & --   & \xmark & \xmark & \xmark & 0.45M \\
Pubmed    & 2008 & \xmark & \xmark & \xmark & 1.10M \\
DBLP      & 2008 & \xmark & \xmark & \xmark & 1.80M \\
S2AG      & 2022  & \checkmark & \checkmark &  \xmark & 205M \\
Aminer    & 2023 & \xmark & \xmark & \xmark & 5.25M \\
Bio-Sci   & 2023 & \xmark & \checkmark & \xmark & 0.03M \\
\midrule
\textbf{Ours} & 2026 & \checkmark & \checkmark$^\dagger$ & \checkmark & 2.0M \\
\bottomrule
\end{tabular}}
\end{wraptable}

\paragraph{Comparison with Existing Datasets.} Table \ref{tab:dataset_comparison} situates our benchmark in the landscape of scientific paper datasets. Existing resources each contribute valuable aspects: APS \citep{bai2019predicting} and DBLP \citep{zhu2018citation} provide structured networks but lack multimodality; PubMed \citep{fu2008models} offers scale but limited label granularity; S2AG \footnote{\url{https://www.semanticscholar.org/}} achieves massive coverage but with sparse annotations; Aminer \footnote{\url{https://www.kaggle.com/datasets/kmader/aminer-academic-citation-dataset}} and Bio-Sci \citep{zhang-etal-2023-content} capture recent slices but often full content. Our dataset is, to our knowledge, the first at this scale (2.0M query papers, 39.3M-node ego-graph) to unify multi-grained impact labels, real author/venue/institution metadata, and built-in processing tools within a single resource -- while being explicit, rather than silent, about the domain shift an unstratified sample of this kind inevitably carries (see Table~\ref{tab:dataset_stats}).

\subsection{Metadata Statistics}
\label{ax:metadata_stats}

\paragraph{Source database.} All papers, authors, venues, affiliations, and citation edges are drawn from a single Semantic Scholar Academic Graph (S2AG) snapshot -- a 307GB database of 237{,}026{,}483 papers spanning 1970--2025. Citation structure comes from the \texttt{citations\_by\_cited}/\texttt{citations\_by\_citing} edge tables (5.7B rows each; every real citation edge is stored twice in this snapshot, confirmed against \texttt{papers.citationcount} and the live Semantic Scholar API, so every query in our pipeline de-duplicates on \texttt{(citing\_corpusid, cited\_corpusid)} before counting), authorship from \texttt{paper\_authors} (697.7M raw rows), and affiliations from \texttt{paper\_affiliations}. We additionally cross-checked S2AG's own \texttt{fields\_of\_study} classification coverage over time: it holds around 65\% for papers published before 2021 and drops to roughly 32\% from 2021 onward without recovering in this snapshot, which is one reason (together with needing a full 5-year label window) that our query window stops at 2020.


\paragraph{Domain distribution.} Table~\ref{tab:full_discipline_ranking} shows the top-15 fields-of-study by share within our 2{,}000{,}000-query sample (a paper may carry more than one field tag, so shares do not sum to 100\%). The distribution is \emph{not} uniform: Medicine alone tags 75.6\% of query papers, roughly triple the next-largest field (Biology, 26.6\%), and coverage falls off from there through Computer Science (21.2\%), Engineering (20.9\%), down to Education (3.5\%) at 15th. This is an honest consequence of unstratified sampling over S2AG's own field mix, not a sampling target -- Table~\ref{tab:corpus_comparison}.

\begin{table}[t]
\centering
\caption{Full corpus vs.\ sampled query set, computed over \texttt{paper\_static\_features} (full corpus) and the union of \texttt{splits.parquet}/\texttt{node\_features.parquet} (sample). Field shares are multi-label and computed independently within each population; ranks refer to each population's own top-15 by share. Full 15-field ranking in Appendix~\ref{ax:dataset_details}, Table~\ref{tab:full_discipline_ranking}.}
\label{tab:corpus_comparison}
\small
\begin{tabular}{l|cc}
\toprule
Attribute & Full S2AG corpus & Sampled query set \\
\midrule
\# Papers & 237{,}026{,}483 & 2{,}000{,}000 (0.84\%) \\
Publication years & 1970--2025 & 2005--2020 \\
Zero-citation share & 53.4\% & 13.8\% \\
5{,}000+-citation share & 0.003--0.008\% & 0.003--0.008\% \\
Physics (share, rank) & 9.7\%, 8th & 21.4\%, 3rd \\
History / Political Science in top-15 & \checkmark\ (14th, 11th) & \xmark\ (dropped) \\
\bottomrule
\end{tabular}
\end{table}

\begin{table}[h]
\centering
\caption{Top-15 field-of-study ranking, full S2AG corpus vs.\ our sampled query set. History and Political Science (full-corpus ranks 14 and 11) fall out of the sample's top-15 entirely, replaced by Agricultural and Food Sciences and Education -- consistent with the referencecount/abstract eligibility filter retaining STEM papers at a higher rate than humanities/social-science papers.}
\label{tab:full_discipline_ranking}
\small
\begin{tabular}{c ll}
\toprule
Rank & Full S2AG corpus (share) & Sampled query set (share) \\
\midrule
1  & Medicine (42.9\%)              & Medicine (75.6\%) \\
2  & Engineering (15.8\%)           & Biology (26.6\%) \\
3  & Computer Science (12.7\%)      & Physics (21.4\%) \\
4  & Biology (12.5\%)               & Computer Science (21.2\%) \\
5  & Chemistry (11.2\%)             & Engineering (20.9\%) \\
6  & Environmental Science (10.7\%) & Environmental Science (18.6\%) \\
7  & Materials Science (9.9\%)      & Materials Science (16.3\%) \\
8  & Physics (9.7\%)                & Chemistry (14.2\%) \\
9  & Psychology (5.7\%)             & Mathematics (12.1\%) \\
10 & Business (4.4\%)               & Psychology (8.6\%) \\
11 & Political Science (4.4\%)      & Economics (5.0\%) \\
12 & Mathematics (4.3\%)            & Sociology (4.9\%) \\
13 & Sociology (4.0\%)              & Business (4.8\%) \\
14 & History (3.9\%)                & Agricultural and Food Sciences (3.6\%) \\
15 & Economics (3.7\%)              & Education (3.5\%) \\
\bottomrule
\end{tabular}
\end{table}

\begin{wrapfigure}{r}{0.5\columnwidth}
  \centering
  \includegraphics[width=\linewidth]{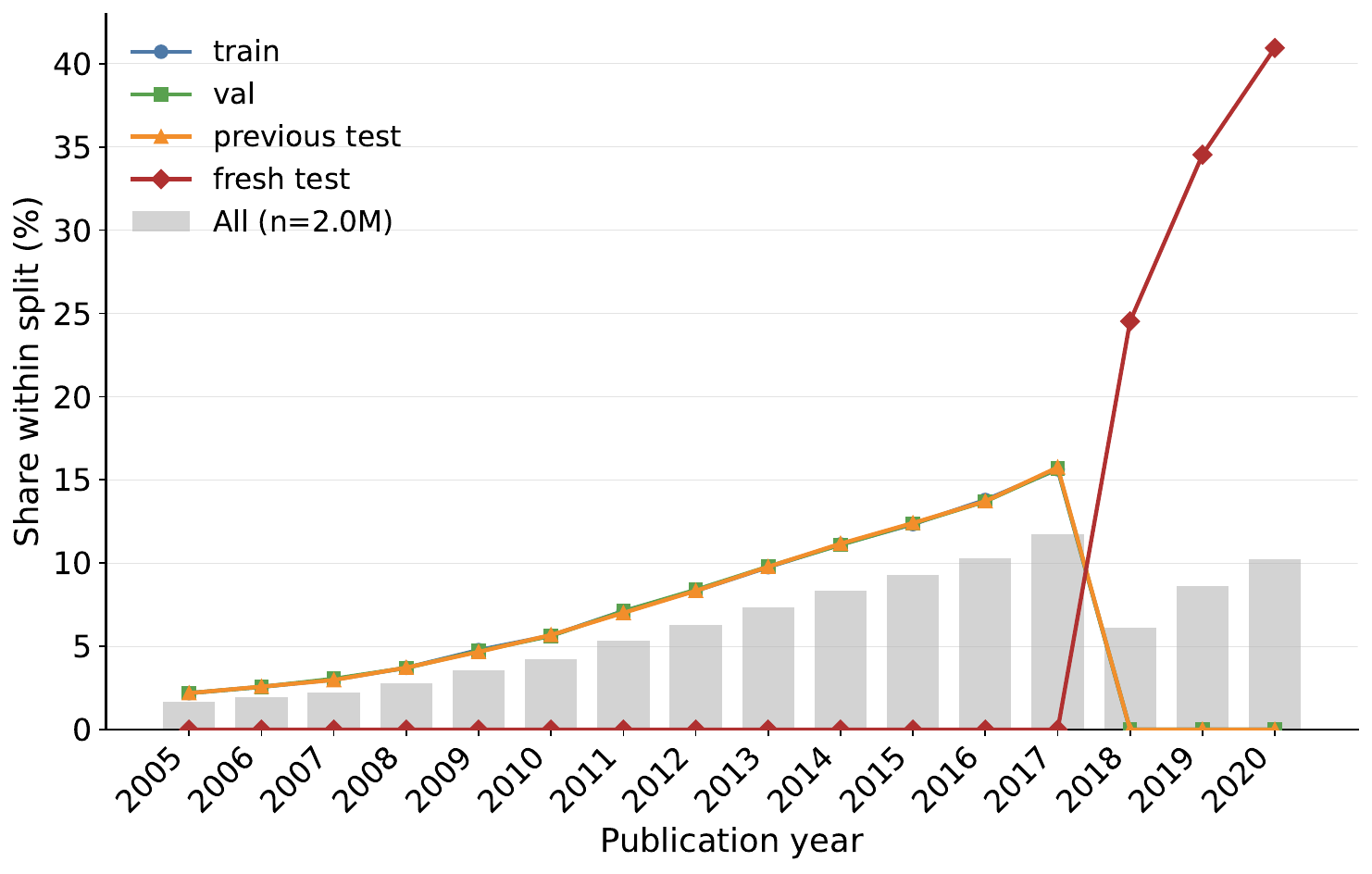}
  \caption{Publication-year share of the full query set (gray bars) and of each split (colored lines). \texttt{train}/\texttt{val}/\texttt{previous\_test} overlap almost exactly (same 2005--2017 pool, same random partition); \texttt{fresh\_test} occupies a disjoint 2018--2020 block.}
  \label{ax_fig:year_distribution}
\end{wrapfigure}

\paragraph{Year distribution, overall and by split.} Figure~\ref{ax_fig:year_distribution} shows the publication-year share of the full 2M-query set (gray bars) against each individual split (colored lines). Because \texttt{train}, \texttt{val}, and \texttt{previous\_test} are all drawn without overlap from the same 2005--2017 historical pool via the same random partition, their year distributions are visually indistinguishable from one another -- all three trace the same growing-then-plateauing curve, rising from 2.1--2.2\% of the split in 2005 to a peak of 15.6--15.9\% in 2017 (more eligible papers survive the referencecount/abstract filter in later years). \texttt{fresh\_test}, by contrast, is drawn entirely from 2018--2020 and is a disjoint block in publication-year space by construction, not merely by chance -- confirming the split does what Section~\ref{sec:dataset_construction} claims: \texttt{fresh\_test} contains publication years the model never saw during training.

\subsection{Split Dataset Distribution}
\label{ax:split_dist}

\begin{figure}[ht]
  \centering
  \includegraphics[width=0.78\linewidth]{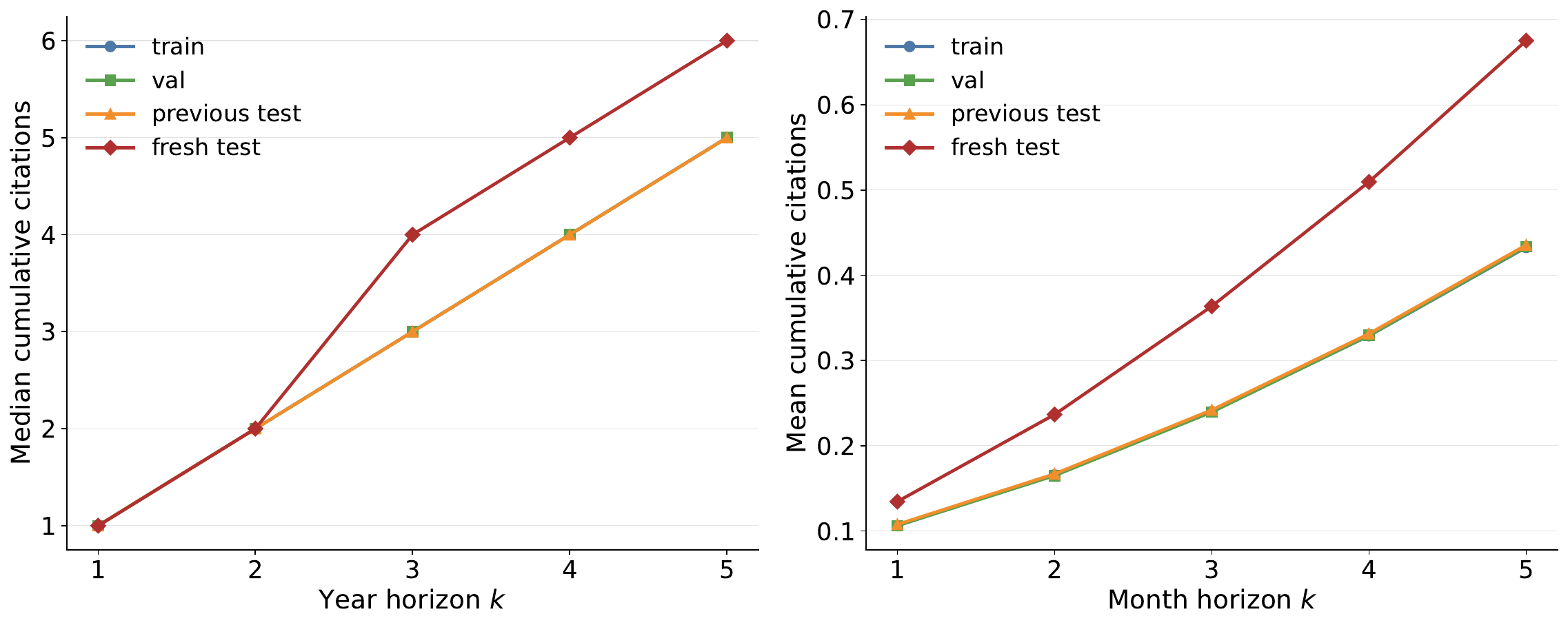}
  \caption{Golden-label citation trajectories by split. \textbf{Left:} median cumulative year-level citations at each horizon. \textbf{Right:} mean cumulative month-level citations at each horizon (mean used since the month-level median is 0 everywhere, per Figure~\ref{fig:year_month_boxplot}). \texttt{train}/\texttt{val}/\texttt{previous\_test} overlap almost exactly; \texttt{fresh\_test} is systematically higher at every horizon and both granularities.}
  \label{ax_fig:citation_golden_distribution}
\end{figure}

Figure~\ref{ax_fig:citation_golden_distribution} compares the golden citation labels themselves across splits: the left panel plots the median cumulative year-level citation count at each horizon $k{=}1..5$, and the right panel the mean cumulative month-level count (mean rather than median, since Figure~\ref{fig:year_month_boxplot} already shows the month-level median is 0 at every horizon). \texttt{train}, \texttt{val}, and \texttt{previous\_test} again track each other almost exactly at both granularities (e.g.\ year-5 median of 5.0 citations, month-5 mean of 0.43), consistent with being three random draws from the same underlying pool. \texttt{fresh\_test} sits systematically higher at every horizon and every granularity -- year-5 median 6.0 vs.\ 5.0, month-5 mean 0.68 vs.\ 0.43 -- which is expected rather than a sampling artifact: 2018--2020 papers are, on average, in more actively-growing subfields of a corpus whose overall publication volume keeps rising, so a same-length citation window started more recently tends to land at a higher point on that growth curve. This also means a model trained on \texttt{train} (drawn from the older, lower-volume regime) faces a real, systematic level shift when evaluated on \texttt{fresh\_test} -- not just a different time window, but a different citation-rate regime -- which is precisely the generalization gap \texttt{fresh\_test} is designed to expose.

\section{Loss Function Details}
\label{ax:method_details}

The composite objective in Eq.~\ref{eq:objective} combines a primary
regression loss with three auxiliary terms, each targeting a different part
of the architecture.

\paragraph{Regression loss $\mathcal{L}_{\mathrm{reg}}$.}
Given the fused, calibrated prediction $\hat{y} \in \mathbb{R}^{T}$ for the
$T$ future citation horizons and the ground-truth cumulative citation counts
$y$, we minimize mean squared error in $\log(1+x)$ space:
\begin{equation}
\mathcal{L}_{\mathrm{reg}} = \frac{1}{T}\sum_{t=1}^{T}\left(\hat{y}_t - \log(1+y_t)\right)^2,
\end{equation}
averaged over the batch. When \texttt{--group\_reweight} is enabled (off by
default), this per-sample MSE is instead combined with learned
domain- and time-specific log-variance terms
$s_d = \texttt{domain\_log\_var}[d]$, $s_\tau = \texttt{time\_log\_var}[\tau]$
in a Kendall-style homoscedastic uncertainty
weighting,
$\mathcal{L}_{\mathrm{reg}} = e^{-(s_d+s_\tau)}\cdot\text{MSE} + s_d + s_\tau$,
so the model can automatically down-weight noisy domain/time slices; we did
not enable this by default and report results with the plain MSE above.

\paragraph{Citation-bucket auxiliary loss $\mathcal{L}_{\mathrm{bucket}}$.}
Each paper is assigned one of three coarse citation-impact buckets based on
its 5-year cumulative citation count $c_5$: \emph{low} ($c_5 < 10$),
\emph{medium} ($10 \le c_5 < 100$), and \emph{high} ($c_5 \ge 100$). A linear
classification head reads the fused branch embedding and predicts bucket
logits $\ell \in \mathbb{R}^3$; $\mathcal{L}_{\mathrm{bucket}}$ is the
standard cross-entropy against the ground-truth bucket label. This is an
auxiliary supervision signal only -- it does not participate in the final
citation-count prediction -- intended to encourage the fused representation
to also be linearly separable by coarse impact tier.

\paragraph{Contrastive loss $\mathcal{L}_{\mathrm{contrast}}$.}
To encourage the graph branch's pooled embedding to be robust to
perturbations of the citation ego-graph's neighborhood, we apply a
hard-negative InfoNCE loss following
H2CGL~\citep{he_h2cgl_2023}. For each anchor graph we construct two augmented
views by randomly dropping a fraction $\rho$ (\texttt{--aug\_rate}, default
$0.1$) of droppable (non-target, non-venue) neighbor nodes, biased toward
dropping low-popularity nodes for the ``positive'' view $z_1$ and
high-popularity nodes for the mined-hard-negative view; batch-mined hard
negatives $z^{-}$ (when available) are pooled with the in-batch negatives.
With $\mathrm{sim}(u,v) = u^\top v / \tau$ ($\tau=0.2$) computed against the
pool of all positives and negatives, excluding the anchor itself:
\begin{equation}
\mathcal{L}_{\mathrm{contrast}} = -\log
\frac{\exp(\mathrm{sim}(z_1, z_2))}
{\sum_{v \in \{z_2\} \cup \mathcal{N} \cup \{z^-\}} \exp(\mathrm{sim}(z_1, v))},
\end{equation}
where $\mathcal{N}$ is the set of in-batch negatives. On \emph{full\_dataset}
we do not yet have a mined hard-negative pool, so $\lambda_{\mathrm{cl}}=0$
there and this term reduces to a standard in-batch InfoNCE (or is fully
disabled); on \emph{calib\_50k} the full hard-negative-augmented version is
used.

\paragraph{Branch deep-supervision loss $\mathcal{L}_{\mathrm{branch}}$.}
Each active branch (LLM, graph, temporal, retrieval, track-record, explicit-
feature, as configured) additionally predicts its own citation estimate
$\hat{y}^{(b)}$ from its own embedding via a lightweight branch-specific
head, independent of the gated fusion. We average the per-branch MSE (same
$\log(1+x)$ target as $\mathcal{L}_{\mathrm{reg}}$) over the $B$ active
branches:
\begin{equation}
\mathcal{L}_{\mathrm{branch}} = \frac{1}{B}\sum_{b=1}^{B}
\frac{1}{T}\sum_{t=1}^{T}\left(\hat{y}^{(b)}_t - \log(1+y_t)\right)^2.
\end{equation}
This keeps every branch directly regression-supervised even though only the
fused prediction is the model's primary output, preventing a branch from
degenerating into a component the gate can freely ignore during early
training.

\paragraph{Weights and training settings.} We set
$\lambda_{\mathrm{aux}}=0.5$ (\texttt{--aux\_weight}),
$\lambda_{\mathrm{cl}}=0.5$ (\texttt{--cl\_weight}, $0$ on
\emph{full\_dataset}), and $\lambda_{\mathrm{branch}}=0.2$
(\texttt{--branch\_aux\_weight}) for all reported results unless otherwise
noted as an ablation.

\section{Evaluation Metrics Definition}
\label{ax:metric_definition}

We report RMSE for year-level prediction and AUROC, PR-AUC, QWK, RMSE,
and NDCG@20 for month-level prediction in our main comparison tables
(Table~\ref{tab:main_previous} and Table~\ref{tab:main_fresh}).
Year-level prediction focuses on regression accuracy, while the
month-level task additionally considers zero-inflation-aware
classification and ranking metrics. Following prior work on citation
forecasting \citep{he_h2cgl_2023,xue2024predicting}, all regression
targets are evaluated in $\log(1+x)$ space to reduce the influence of the
heavy right tail of citation-count distributions. We consider the same
set of $H=5$ future citation horizons for both prediction granularities:
years 1--5 for the \emph{Year} target and months 1--5 for the
\emph{Month} target.

\paragraph{Notation.}
Let $\mathcal{D}_{\text{test}} = \{1,\ldots,N\}$ denote a test split,
such as \textsc{previous\_test} or \textsc{fresh\_test}. Let
$c_{i,h}$ denote the raw cumulative citation count of paper $i$ at
horizon $h \in \{1,\ldots,H\}$. We evaluate regression metrics in
$\log(1+x)$ space:
\begin{equation}
    y_{i,h} = \log(1+c_{i,h}),
    \qquad
    \hat{y}_{i,h} = \log(1+\hat{c}_{i,h}),
\end{equation}
where $\hat{c}_{i,h}$ is the predicted cumulative citation count.
For ranking and regression metrics, the transformed values are used
consistently throughout evaluation.

\subsection{Year-Level Metrics}

\paragraph{RMSE.}
We report per-horizon RMSE,
\begin{equation}
    \mathrm{RMSE}_h
    =
    \sqrt{
        \frac{1}{N}
        \sum_{i=1}^{N}
        \left(
            \hat{y}_{i,h}-y_{i,h}
        \right)^2
    },
\end{equation}
and define the pooled RMSE across all $N\times H$ paper--horizon pairs as
\begin{equation}
    \mathrm{RMSE}_{\mathrm{overall}}
    =
    \sqrt{
        \frac{1}{NH}
        \sum_{i=1}^{N}
        \sum_{h=1}^{H}
        \left(
            \hat{y}_{i,h}-y_{i,h}
        \right)^2
    }.
\end{equation}
The \emph{Year Avg} and \emph{Month Avg} columns in
Table~\ref{tab:main_previous} and Table~\ref{tab:main_fresh} report the
unweighted mean of the per-horizon RMSE values,
\begin{equation}
    \mathrm{RMSE}_{\mathrm{Avg}}
    =
    \frac{1}{H}
    \sum_{h=1}^{H}
    \mathrm{RMSE}_h.
\end{equation}
This treats all prediction horizons equally and prevents horizons with
larger residual variance from dominating the aggregate score.

\paragraph{NDCG@20.}
RMSE and classification metrics do not directly measure whether a model
correctly ranks papers by future citation impact. We therefore report
NDCG@20 to evaluate the quality of the highest-ranked papers. For each
horizon $h$, we treat the entire test split as a single ranked list,
sorting papers according to the model's predicted scores
$\hat{y}_{\cdot,h}$. The discounted cumulative gain at rank $k$ is
\begin{equation}
    \mathrm{DCG}_h@k
    =
    \sum_{r=1}^{k}
    \frac{2^{\mathrm{rel}(r)}-1}
         {\log_2(r+1)},
\end{equation}
where $\mathrm{rel}(r)$ is the ground-truth relevance of the paper
ranked at position $r$, here given by $y_{i,h}$. We set $k=20$ and
normalize by the ideal DCG obtained by ranking papers according to
ground-truth relevance:
\begin{equation}
    \mathrm{NDCG}_h@20
    =
    \frac{
        \mathrm{DCG}_h@20
    }{
        \mathrm{IDCG}_h@20
    }.
\end{equation}
We report NDCG@20 as the mean of the five per-horizon values. Higher
NDCG@20 indicates better agreement between the model's top-20 predicted
ranking and the ranking induced by realized citation impact.

NDCG@20 is computed globally over the full evaluation split rather than
independently within queries, batches, or distributed evaluation shards.
When evaluation is parallelized across multiple devices, we first
reassemble the predictions and ground-truth labels from all shards and
then compute NDCG@20 once over the complete split. This avoids averaging
rank-based scores computed on disjoint subsets, which would not preserve
the ranking relationships in the full evaluation set.

\subsection{Month-Level Metrics}

\paragraph{AUROC and PR-AUC.}
Month-level citation targets are highly zero-inflated, with 85.2\% of
the valid previous-test labels being zero. We therefore complement
regression with binary discrimination metrics. For each horizon $h$, we
define the binary citation indicator
\begin{equation}
    z_{i,h} = \mathbb{I}[c_{i,h} > 0],
\end{equation}
where $z_{i,h}=1$ indicates that the paper receives at least one citation
by horizon $h$. The model's predicted citation score is used as the
continuous decision score for both metrics.

AUROC measures the probability that a randomly selected cited paper
receives a higher predicted score than a randomly selected uncited
paper. Equivalently, it is the area under the receiver operating
characteristic curve obtained by varying the decision threshold over
the predicted scores. PR-AUC is the area under the precision--recall
curve and is particularly informative under severe class imbalance.
We report the corresponding scores for each month-level horizon and
aggregate them consistently across the five horizons.

\paragraph{Quadratic Weighted Kappa (QWK).}
To evaluate ordinal agreement between predicted and realized citation
impact, we discretize cumulative citation counts into three ordered
impact buckets, \emph{low}, \emph{medium}, and \emph{high}, using fixed
citation-count thresholds. The same thresholds are applied to all
models and splits. Let $q_{i,h}$ and $\hat{q}_{i,h}$ denote the ground
truth and predicted ordinal categories, respectively. QWK measures the
agreement between the two ordinal labelings while assigning larger
penalties to larger disagreements. Following the standard formulation,
\begin{equation}
    \mathrm{QWK}
    =
    1 -
    \frac{
        \sum_{a,b} w_{ab} O_{ab}
    }{
        \sum_{a,b} w_{ab} E_{ab}
    },
\end{equation}
where $O_{ab}$ is the observed confusion matrix, $E_{ab}$ is the
expected confusion matrix under the empirical marginal distributions,
and
\begin{equation}
    w_{ab}
    =
    \frac{(a-b)^2}{(K-1)^2},
    \qquad K=3.
\end{equation}
Higher QWK indicates stronger ordinal agreement between predicted and
ground-truth citation-impact levels.

\paragraph{Splits.}
We report the evaluation metrics separately on
\textsc{previous\_test} and \textsc{fresh\_test}. The
\textsc{previous\_test} split contains papers whose publication years
overlap with the training period and therefore evaluates
in-distribution temporal generalization. In contrast,
\textsc{fresh\_test} contains papers published strictly after the
training cutoff and evaluates extrapolation under temporal distribution
shift. This separation allows us to distinguish predictive performance
within the historical data distribution from robustness to unseen
publication periods.

\section{More Experimental Results}

\subsection{Significance Test}

We assess the statistical significance of the performance differences
between \textit{LLM4Impact} and its ablations using a paired bootstrap test. Predictions are matched by \texttt{corpusid}, ensuring that each comparison is performed on exactly the same set of papers. As shown in Table \ref{tab:bootstrap_significance}, we draw 2,000 bootstrap resamples and
report the RMSE difference, defined as the ablation RMSE minus the RMSE of
\textit{LLM4Impact}, together with its 95\% confidence interval and $p$-value. All comparisons yield $p < 0.0001$, with confidence intervals entirely above
zero, confirming that the observed improvements of \textit{LLM4Impact} over the
ablated variants are statistically significant on both the previous and
fresh test sets.

\begin{table}[t]
\centering
\caption{Paired bootstrap significance tests comparing
\textit{LLM4Impact} with its ablations.
RMSE differences are computed as
$\mathrm{RMSE}_{\mathrm{abl.}}-\mathrm{RMSE}_{\textit{LLM4Impact}}$.
Positive values indicate that the ablation performs worse than
\textit{LLM4Impact}.}
\label{tab:bootstrap_significance}
\resizebox{0.7\linewidth}{!}{%
\begin{tabular}{llccc}
\toprule
Ablation & Split & RMSE Diff. & 95\% CI & $p$ \\
\midrule
\texttt{no\_graph}
    & \texttt{previous\_test} & $+0.0482$ & $[0.0467,\,0.0496]$ & $<0.0001$ \\
\texttt{no\_llm}
    & \texttt{previous\_test} & $+0.0891$ & $[0.0872,\,0.0910]$ & $<0.0001$ \\
\texttt{no\_llm\_graph\_prefix}
    & \texttt{previous\_test} & $+0.0033$ & $[0.0023,\,0.0043]$ & $<0.0001$ \\
\texttt{no\_graph}
    & \texttt{fresh\_test} & $+0.0328$ & $[0.0312,\,0.0343]$ & $<0.0001$ \\
\texttt{no\_llm}
    & \texttt{fresh\_test} & $+0.1028$ & $[0.1010,\,0.1046]$ & $<0.0001$ \\
\bottomrule
\end{tabular}%
}
\end{table}

\subsection{More Case Study}
\label{ax:more_case_study}

\begin{figure}[t]
    \centering
    \includegraphics[width=0.99\textwidth]{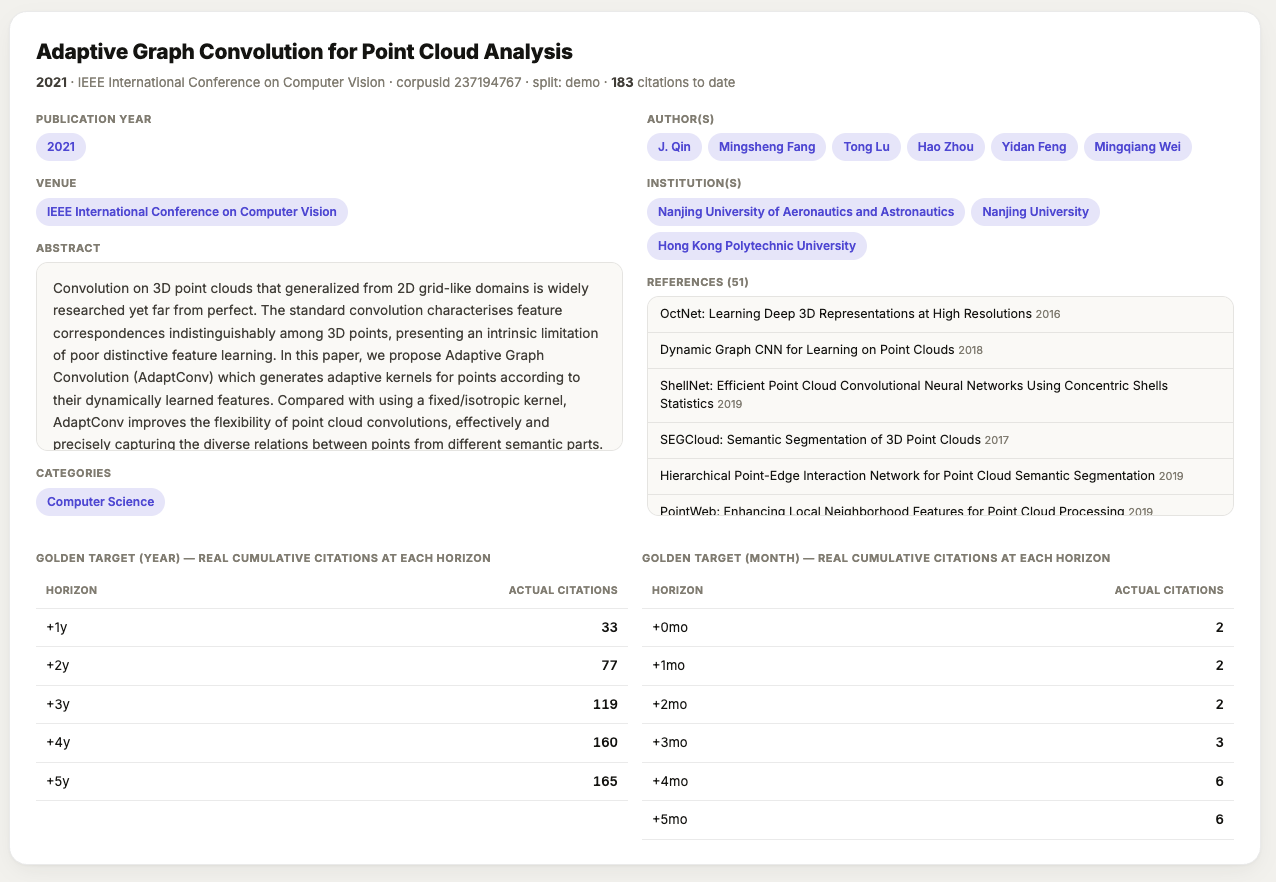}
    \caption{Example benchmark instance and data format, including paper
metadata, references, and ground-truth citation targets at year and
month granularities.}
    \label{fig:ax_dataset_example}
\end{figure}

\begin{figure}[t]
    \centering
    \includegraphics[width=0.99\textwidth]{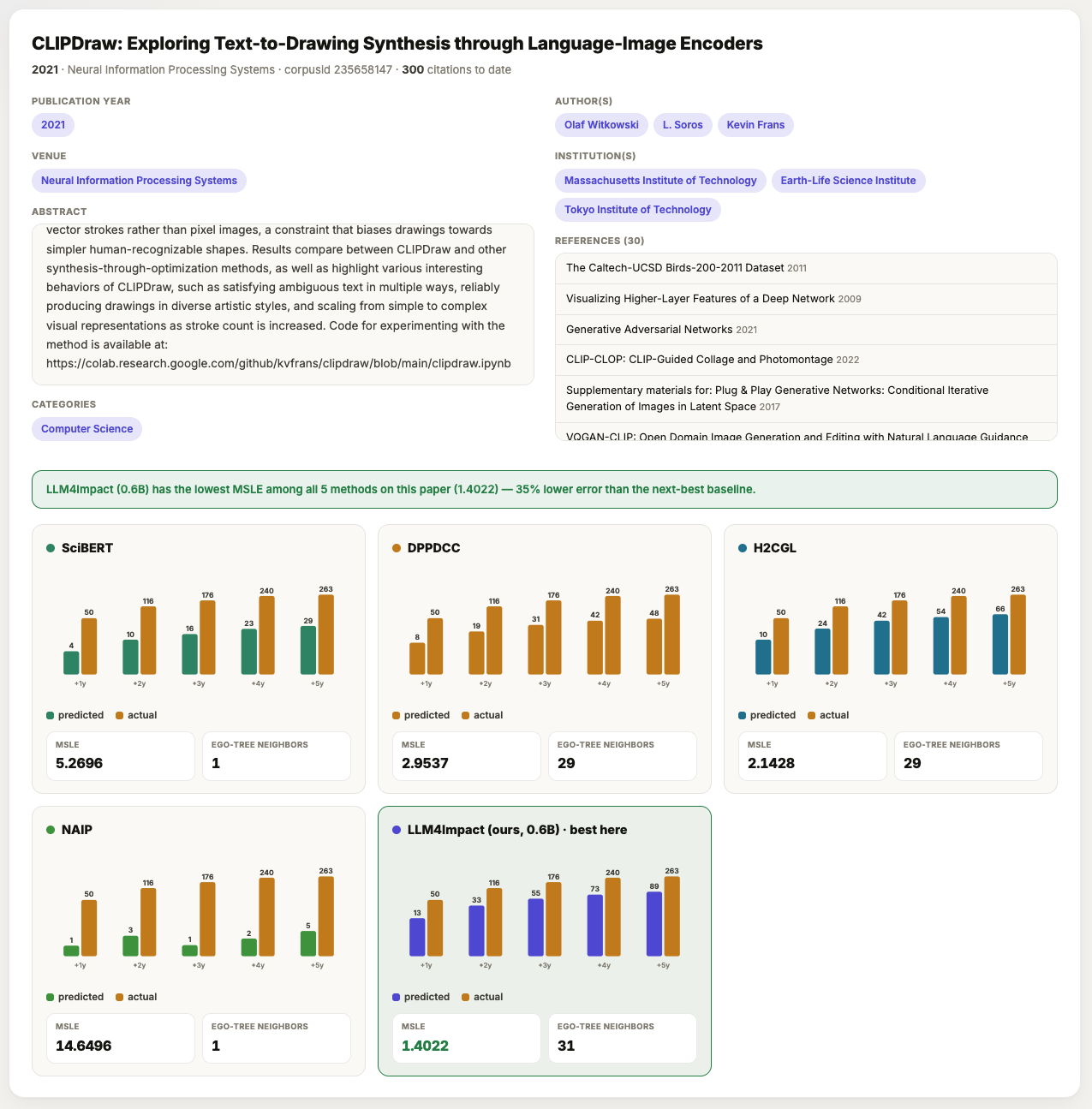}
    \caption{Interactive case study comparing LLM4Impact with baseline
models on a real paper, showing predicted and ground-truth citation
trajectories.}
    \label{fig:ax_case_1}
\end{figure}

\begin{figure}[t]
    \centering
    \includegraphics[width=0.99\textwidth]{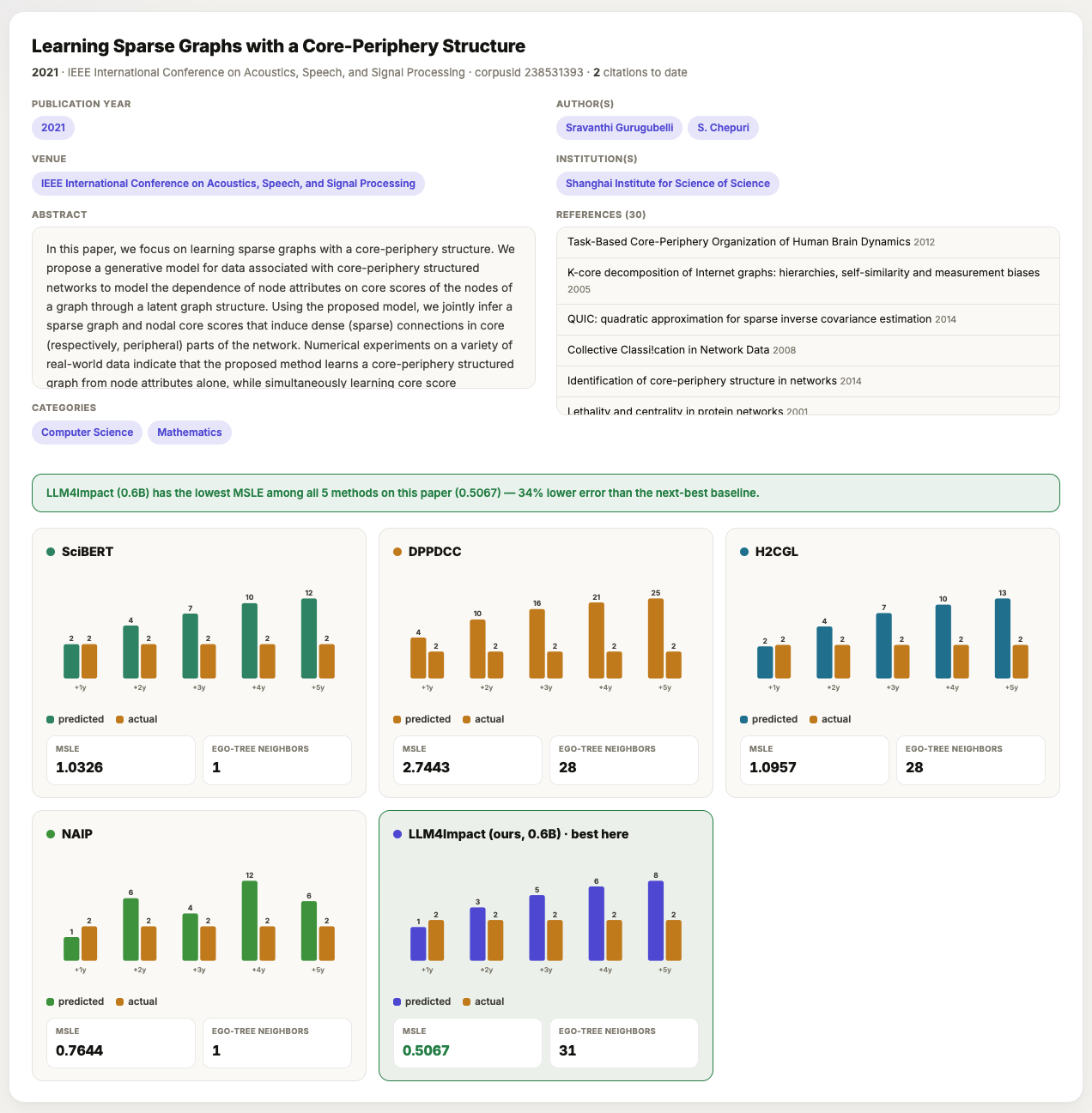}
    \caption{Second interactive case study comparing LLM4Impact with
baseline models on a real paper, showing predicted and ground-truth
citation trajectories.}
    \label{fig:ax_case_2}
\end{figure}

\begin{figure}[t]
    \centering
    \includegraphics[width=0.99\textwidth]{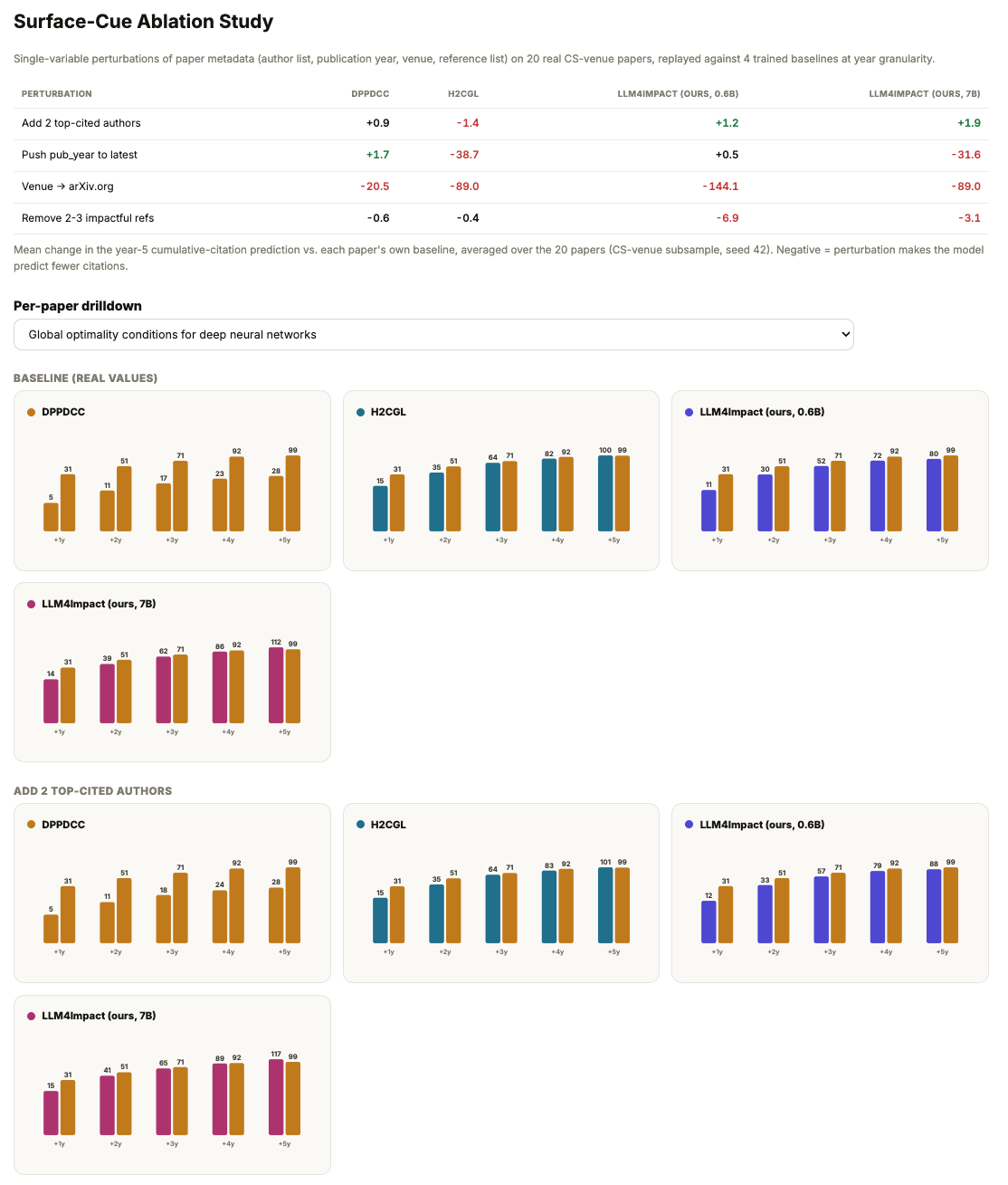}
    \caption{Surface-cue perturbation study examining the effects of
author, publication-year, venue, and reference changes.}
    \label{fig:ax_control_exp}
\end{figure}

\begin{figure}[t]
    \centering
    \includegraphics[width=0.99\textwidth]{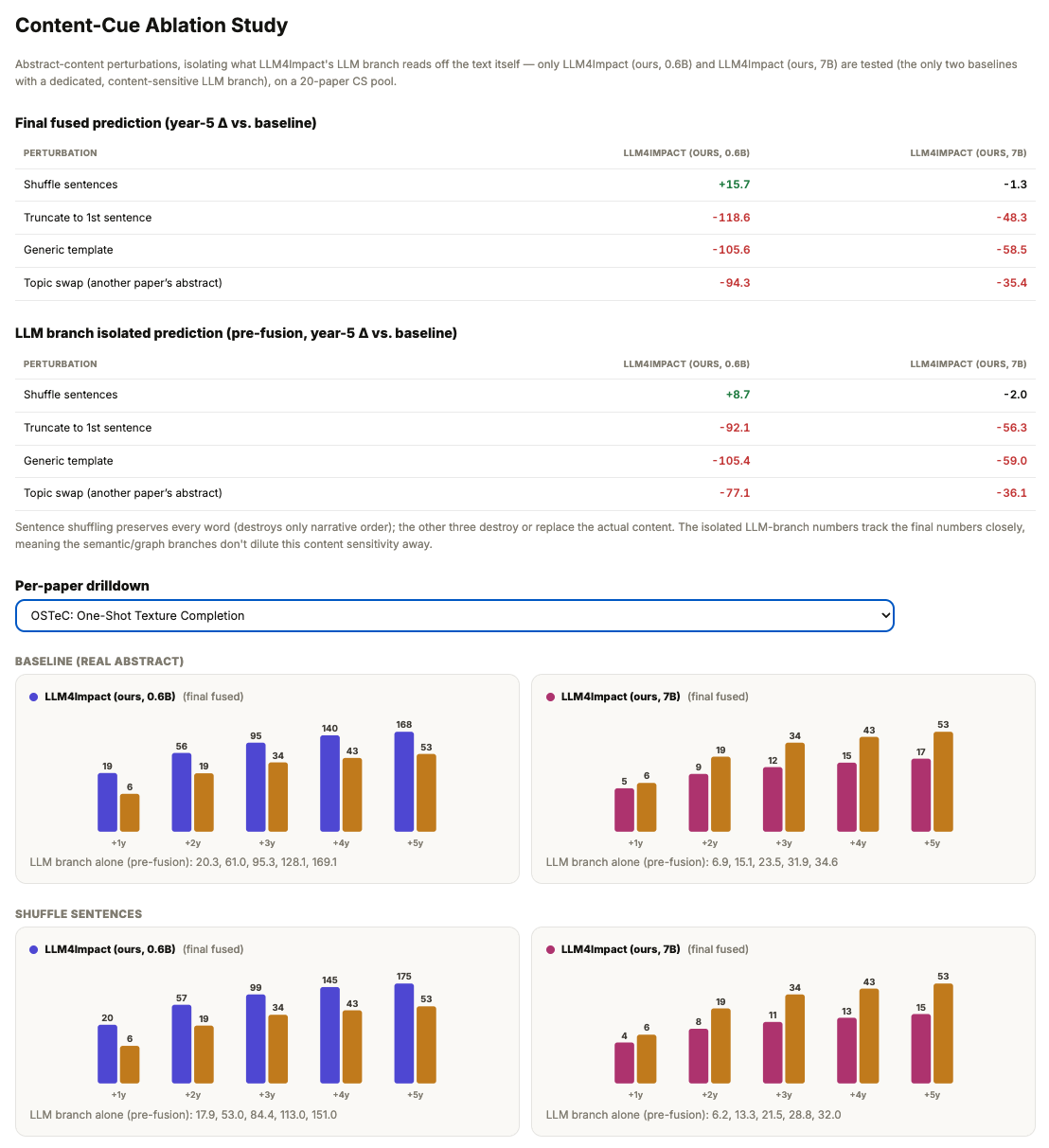}
    \caption{Content-cue perturbation study illustrating the role of the
LLM branch under controlled abstract perturbations.}
    \label{fig:ax_llm_usage}
\end{figure}

To facilitate qualitative inspection of our method's behavior, we accompany this submission
  with an interactive demo website (included in the supplementary material) that allows a
  reader to explore model predictions on real papers without running any code. The demo
  provides four components, illustrated in Figures~\ref{fig:ax_dataset_example}--\ref{fig:ax_llm_usage}:

  \begin{itemize}
      \item \textbf{Case Gallery} (Figure~\ref{fig:ax_dataset_example}): a curated set of real papers
      from top CS/ML venues, each showing our method's predicted citation trajectory alongside
      five baselines and the actual (golden) citation counts, together with the paper's full
      metadata (title, venue, publication year, authors, institutions, categories, abstract,
      and references). Cases are tagged as wins, typical, or losses for our method, so the
      gallery is not restricted to favorable examples.
      \item \textbf{Dataset Examples} (Figure~\ref{fig:ax_case_1} and \ref{fig:ax_case_2}): a browsable collection
      of papers from the evaluation pool with complete metadata and their ground-truth citation
      trajectories at both year and month granularity, illustrating the task itself independent
      of any model's predictions.
      \item \textbf{Surface-Cue Ablation Study} (Figure~\ref{fig:ax_control_exp}): an
      interactive view of the single-variable metadata perturbation study described in
      Section~\ref{sec:more_analysis} (author, publication year, venue, and reference perturbations),
      with per-paper drill-down.
      \item \textbf{Content-Cue Ablation Study} (Figure~\ref{fig:ax_llm_usage}): an
      interactive view of the abstract-content perturbation study, isolating what our LLM
      branch reads from the paper text itself.
  \end{itemize}

We sincerely hope readers to consult the submitted website
  directly for additional examples beyond what is shown here.

\subsection{Domain Performance Differences}

\begin{figure}[t]
\centering
\includegraphics[width=\linewidth]{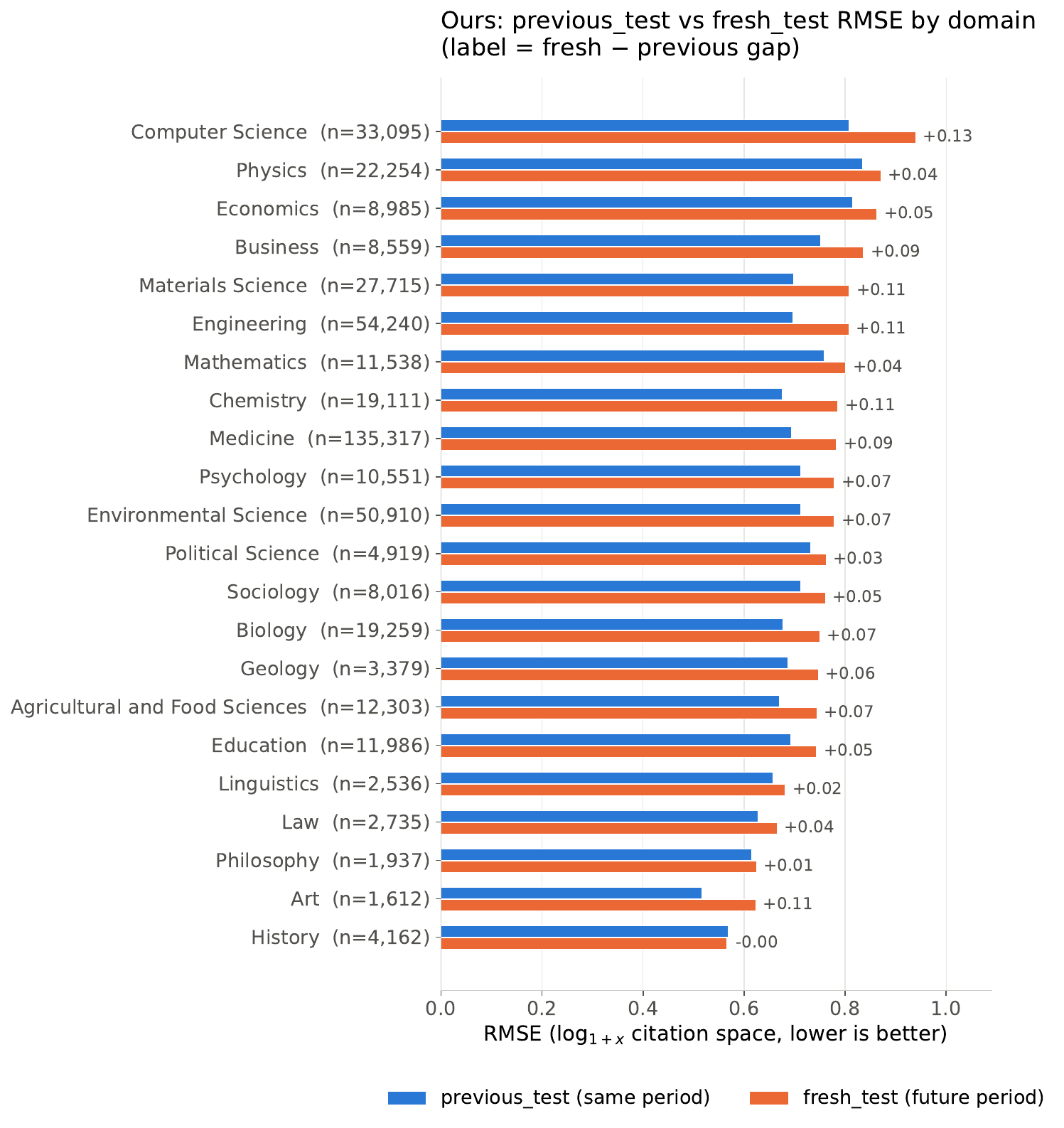}
\caption{RMSE of the full model on \emph{previous\_test} (same time period
as training) versus \emph{fresh\_test} (held-out future time period),
broken down by primary field of study. Domains are sorted by
\emph{fresh\_test} RMSE (easiest at bottom). The label at the end of each
domain's bars is the fresh$-$previous gap, i.e.\ how much harder
extrapolating to a future time period is for that field.}
\label{fig:domain_split_gap}
\end{figure}

Figure~\ref{fig:domain_split_gap} breaks down the previous\_test/fresh\_test
gap by primary field of study. The gap is not uniform across domains:
\textbf{Computer Science shows both the largest absolute RMSE and the
largest previous\_test$\to$fresh\_test gap} (+0.13), suggesting its citation
dynamics are both harder to predict overall and less stable across time
periods than other fields. In contrast, \textbf{History has an essentially
zero gap} ($-0.00$), and several small-sample humanities domains (Law,
Philosophy, Linguistics) also show gaps under $+0.05$, consistent with
citation patterns in these fields evolving more slowly and thus
generalizing better to a future test period. This pattern held across all
$22$ domains with $\geq\!1{,}000$ examples in our fresh\_test split.

\section{TNCSI Calculation}
\label{ax:section_ndcg}

TNCSI (Topic Normalized Citation Success Index) is a metric designed to evaluate the citation impact of articles by comparing them within the same research field, providing a normalized score between 0 and 1 that indicates the likelihood an article’s impact surpasses that of its peers. Unlike traditional citation counts, TNCSI normalizes across fields but originally focused on review papers and cumulative citations, limiting its suitability for comparing newly published or regular research articles. The formulation is as follows:

\begin{equation}
    TNCSI = \int_{0}^{\mathrm{cites}} \lambda e^{-\lambda x}, dx,\quad x \ge 0.
\end{equation}

\begin{table}[t]
\centering
\caption{Main hyperparameters for the reported full-dataset results.}
\label{tab:hyperparams}
\resizebox{0.7\textwidth}{!}{
\begin{tabular}{llc}
\toprule
\textbf{Category} & \textbf{Hyperparameter} & \textbf{Value} \\
\midrule
\multirow{5}{*}{Architecture}
& Hidden dimension & 256 \\
& Graph layers / attention heads & 3 / 8 \\
& LLM backbone & Qwen3-0.6B / Qwen2.5-7B-Instruct$^\ddagger$ \\
& LoRA rank $r$ / alpha $\alpha$ / dropout & 64 / 32 / 0.05 \\
& Graph-prefix length & 10 tokens \\
\midrule
\multirow{3}{*}{Loss weights}
& $\lambda_{\mathrm{aux}}$ (bucket classification) & 0.5 \\
& $\lambda_{\mathrm{cl}}$ (contrastive) & 0.5 \\
& $\lambda_{\mathrm{branch}}$ (deep supervision) & 0.2 \\
\midrule
\multirow{6}{*}{Optimization}
& Optimizer & AdamW \\
& Learning rate (constant) & $2\times10^{-5}$ \\
& Weight decay & 0.01 \\
& Per-GPU / effective batch size & 8 / 32 (0.6B), 4 / 16 (7B) \\
& GPUs & 4$\times$A100-40GB \\
& Training length & 1 full epoch \\
\midrule
\multirow{4}{*}{Data}
& Train / val examples & 800K / 300K \\
& previous\_test / fresh\_test examples & 400K / 500K \\
& Prediction horizons $T$ & 5 \\
& Ego-tree hop depth & 1 \\
\bottomrule
\end{tabular}}
\\[2pt]
\end{table}

\section{Training parameters}

We implement all branches in PyTorch. Our main results use \texttt{Qwen3-0.6B} as the frozen LLM backbone; for a same-backbone comparison against NAIP, we additionally report a \texttt{Qwen2.5-7B-Instruct} backbone variant. Both are adapted with LoRA ($r=64$, $\alpha=32$, applied only to the attention $q_{\mathrm{proj}}/v_{\mathrm{proj}}$ projections) rather than full fine-tuning. The graph branch encodes a 1-hop ego-tree around each target paper with a 3-layer, 8-head relational graph attention network, and its pooled embedding is projected into $10$ soft prefix tokens prepended to the LLM's input sequence. We optimize the composite objective in Eq.~\ref{eq:objective} with AdamW at a constant learning rate of $2\times10^{-5}$ (no warmup or decay schedule) and weight decay $0.01$, training for a single full epoch over the 800K-example training set on $4\times$ NVIDIA A100-40GB GPUs under distributed data parallelism, with validation checked every $500$ steps (used only for checkpoint selection, not for early termination -- we always complete the full epoch). Per-GPU batch size is $8$ (effective $32$) for the 0.6B backbone and $4$ (effective $16$) for the 7B backbone, reduced to fit the larger backbone's memory footprint alongside the graph branch. We evaluate on two disjoint test splits: \emph{previous\_test} (400K examples, same time period as training) and \emph{fresh\_test} (500K examples, a held-out future time period), predicting cumulative citation counts at $T=5$ future horizons. Table~\ref{tab:hyperparams} summarizes the main hyperparameters; the auxiliary contrastive loss weight $\lambda_{\mathrm{cl}}$ is set to $0.5$ on the full dataset.

\section{Prompt}

\begin{tcolorbox}[
    colframe=promptbox_color,
    colback=white,
    coltitle=black,
    colbacktitle=promptbox_color,
    rounded corners,
    arc=1.5mm, 
    enhanced,                    
    boxrule=0.6mm,                
    frame style={solid},          
    fonttitle=\bfseries,          
    fontupper=\small,
    title={The prompt of LLMs to predict future citation counts.}
]

You are a scientometrics analyst. The tokens preceding this text encode
structural context about the paper -- its citation graph, retrieved related
papers, and its own venue, institution, authors, and references. Using both
that context and the paper's own content below, predict its cumulative
citation counts at several future time points after publication.
Title: {title}
Abstract: {abstract}

\end{tcolorbox}

\begin{wraptable}{l}{0.5\linewidth}
\small
\caption{Comparison of RMSE for raw citation counts and TNCSI,
together with their Spearman correlation, on the previous test set.
$C_{\log}$ denotes log-transformed citation counts and
$C_{\mathrm{TNCSI}}$ denotes TNCSI-based impact scores.}
\label{tb:comparsion_of_tncsi_previous}
\resizebox{0.5\textwidth}{!}{
\begin{tabular}{l|ccccc|c}
\toprule
Metric   & Year\_1      & Year\_2      & Year\_3      & Year\_4      & Year\_5      & Avg    \\ \midrule
\multicolumn{7}{c}{\textit{Previous}} \\
C\_log   & {0.6772} & {0.6932} & {0.7809} & {0.8022} & {0.9177} & {0.7719}  \\ \midrule
C\_TNCSI & 0.2016 & 0.2211 & 0.2291 & 0.2327 & 0.2352 & 0.2239 \\ \midrule
Cor.     & 0.7900 & 0.8061 & 0.8062 & 0.8119 & 0.8135 & 0.8055 \\ \midrule \midrule
\bottomrule
\end{tabular}}
\end{wraptable}

\clearpage

\begin{tcolorbox}[
    colframe=promptbox_color,
    colback=white,
    coltitle=black,
    colbacktitle=promptbox_color,
    rounded corners,
    arc=1.5mm, 
    enhanced,                    
    boxrule=0.6mm,                
    frame style={solid},          
    fonttitle=\bfseries,          
    fontupper=\small,
    title={The prompt of LLMs to normalize noisy affiliations.}
]
You are given a list of affiliation strings. Some are valid institutions, some are duplicated, and some are meaningless. Please normalize them as follows:

\begin{itemize}
\item If the string is a valid institution name but with formatting issues, fix it (e.g., remove extra punctuation, unify into ``Institution Name'').
\item If the string is a duplicate of another name, map it to the same corrected name.
\item If the string is not an institution name (number, meaningless), map it to ``unknown''.
\item Return the result strictly as a JSON dictionary, where each original string is mapped to its corrected normalized name.
\end{itemize}

\textbf{Example Input:}\\
\text{[}``University of California, Los Angeles'', ``University of California, Los Angeles'', ``01 Collaboration''\text{]}\\

\textbf{Example Output:}

\begin{tcblisting}{listing only, colback=white, colframe=promptbox_color!70, enhanced, boxrule=0.3mm, 
  listing options={breaklines=true}}
{
  "University of California, Los Angeles": "University of California",
  "01 Collaboration": "unknown"
}
\end{tcblisting}

\textbf{Generation Output:}\\
Please normalize the following affiliation list accordingly.
\end{tcolorbox}

\begin{tcolorbox}[
    colframe=promptbox_color,
    colback=white,
    coltitle=black,
    colbacktitle=promptbox_color,
    rounded corners,
    arc=1.5mm, 
    enhanced,                    
    boxrule=0.6mm,                
    frame style={solid},          
    fonttitle=\bfseries,          
    fontupper=\small,
    title={The prompt of LLMs to normalize noisy venues.}
]
You are given a list of venue strings. Some are valid venues, some are duplicated, and some are meaningless. Please normalize them as follows:

\begin{itemize}
\item If the string is a valid venue name but with formatting issues, fix it (e.g., remove extra punctuation, use proper title case like ``Venue Name'').
\item Remove any explicit year tokens (e.g., ``2015'', ``2014'').
\item Remove edition indicators such as ordinal numbers or words (e.g., ``1st'', ``2nd'', ``Third'', ``Fourth'').
\item If the venue contains a trailing parenthetical acronym/abbreviation like ``(ICAC3N)'', ``(HPEC)'', ``(IVAPP)'' that is all caps/digits/hyphen and $\leq$10 chars, drop that parenthetical part. Keep descriptive tracks in parentheses (e.g., ``(Emerging Technologies)'') as is.
\item If both a full spelled-out name and an acronym appear together, keep only the full spelled-out name and drop the acronym.
\item If the string is a duplicate of another (after normalization), map it to the same corrected name.
\item If the string is not a venue name (pure number or meaningless), map it to ``unknown''.
\item Return the result strictly as a JSON dictionary, where each original string is mapped to its corrected normalized name. Do not include comments or extra text.
\end{itemize}

\textbf{Example Input:}\\
\text{[}``Journal of Machine Learning Research'', ``2015 IEEE International Conference on Web Services'', ``4th International Conference on X (ICACX)'', ``01 Collaboration''\text{]}\\

\textbf{Example Output:}

\begin{tcblisting}{listing only, colback=white, colframe=promptbox_color!70, enhanced, boxrule=0.3mm, 
  listing options={breaklines=true}}
{
  "Journal of Machine Learning Research": "Journal of Machine Learning Research",
  "2015 IEEE International Conference on Web Services": "IEEE International Conference on Web Services",
  "4th International Conference on X (ICACX)": "International Conference on X",
  "01 Collaboration": "unknown"
}
\end{tcblisting}

\textbf{Generation Output:}\\
Please normalize the following venue list accordingly.
\end{tcolorbox}

\section{Limitations and Future Work}

Our study focuses on a finite set of scientific domains, publication years, and prediction horizons, leaving broader settings for future investigation. 
In particular, longer-term citation dynamics and the continuous evolution of scholarly networks may provide additional signals beyond our current point-in-time representation. 
Future work will therefore explore broader domains, longer prediction horizons, and dynamic representations of evolving scientific knowledge.

\end{document}